\documentclass[10pt,twocolumn,letterpaper]{article}

\usepackage[pagenumbers]{cvpr} 

\usepackage{microtype}

\definecolor{cvprblue}{rgb}{0.21,0.49,0.74}
\usepackage[pagebackref,breaklinks,colorlinks,allcolors=cvprblue]{hyperref}
\usepackage{array}
\usepackage{multirow}
\usepackage{multicol}
\usepackage{makecell}
\usepackage{textcomp}
\usepackage{upgreek}
\usepackage{amsmath}
\usepackage{mathtools}
\usepackage{booktabs}
\usepackage{bbding}
\newcommand{\myPara}[1]{\noindent\textbf{#1}}
\usepackage[accsupp]{axessibility}  

\def\paperID{34954} 
\def\confName{CVPR}
\def\confYear{2026}

\title{Seeing through Light and Darkness: Sensor-Physics Grounded Deblurring HDR NeRF from Single-Exposure Images and Events}

\author{
Yunshan Qi\textsuperscript{1}\hspace{4ex}
Lin Zhu\textsuperscript{2}\footnotemark[1]\hspace{4ex}
Nan Bao\textsuperscript{1}\hspace{4ex}
Yifan Zhao\textsuperscript{1}\hspace{4ex}
Jia Li\textsuperscript{1}\thanks{Correspondence should be addressed to Jia Li and Lin Zhu. Website: \url{https://cvteam.buaa.edu.cn}}\\[0.7ex]
\textsuperscript{1}State Key Laboratory of Virtual Reality Technology and Systems, SCSE \& QRI, Beihang University \\[0.5ex]
\textsuperscript{2}School of Artificial Intelligence, Beijing Normal University\hspace{2.0ex}\\ [0.6ex]
{\tt\small \{qi\_yunshan, nbao, zhaoyf, jiali\}@buaa.edu.cn,}\hspace{1ex}
{\tt\small linzhu@bnu.edu.cn}\\
}

\begin{document}
\maketitle


\begin{abstract}
Novel view synthesis from low dynamic range (LDR) blurry images, which are common in the wild, struggles to recover high dynamic range (HDR) and sharp 3D representations in extreme lighting conditions.
Although existing methods employ event data to address this issue, they ignore the sensor-physics mismatches between the camera output and physical world radiance, resulting in suboptimal HDR and deblurring results.
To cope with this problem, we propose a unified sensor-physics grounded NeRF framework for sharp HDR novel view synthesis from single-exposure blurry LDR images and corresponding events.
We employ NeRF to directly represent the actual radiance of the 3D scene in the HDR domain and model raw HDR scene rays hitting the sensor pixels as in the physical world.
A 2D pixel-wise RGB CRF model is introduced to align the NeRF rendered pixel values with the sensor-recorded LDR pixel values of the input images.
A novel event CRF model is also designed to bridge the gap between physical scene dynamics and event sensor output.
The two models are jointly optimized with the NeRF network, leveraging the spatial and temporal dynamic information in events to enhance the sharp HDR 3D representation learning.
Experiments on the collected and public datasets demonstrate that our method achieves state-of-the-art HDR and deblurring novel view synthesis results with single-exposure blurry LDR images and corresponding events.
Our code and datasets are publicly available at \url{https://github.com/iCVTEAM/See-NeRF}.
\end{abstract}

\begin{figure}[t]
\centering
\includegraphics[width=1\linewidth]{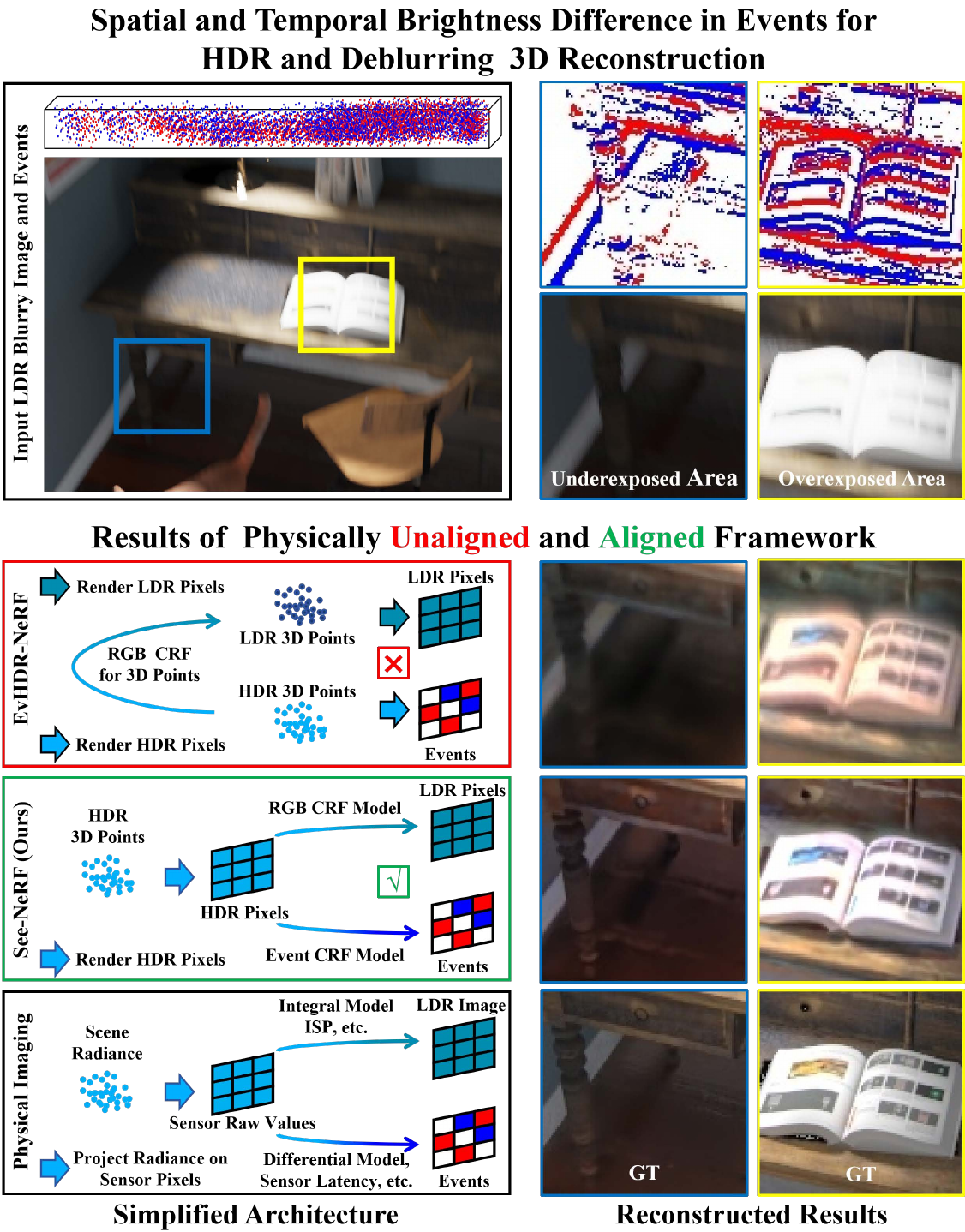}
\caption{Events corresponding to a blurry LDR image contain spatial difference (High Dynamic Range) and temporal difference (High Temporal Resolution) information of the scene radiance, making it an ideal data for enhancing HDR sharp 3D scene reconstruction.
See-NeRF models the physical imaging process with proposed RGB and event CRF models to bridge the sensor-physics discrepancy, leveraging events to infer scene dynamics and achieving sharper and better HDR novel view synthesis results compared to the physical unaligned method EvHDR-NeRF~\cite{Evhdr-NeRF}.}
\label{fig:1}
\vspace{-1em}
\end{figure}


\section{Introduction}
\label{sec:1}

Recent novel view synthesis (NVS) works, Neural Radiance Fields (NeRF)~\cite{nerf} and 3D Gaussian Splatting (3DGS)~\cite{3DGS}, achieve high-fidelity and high-speed NVS results with ideal low dynamic range (LDR) sharp images as input, respectively.
However, their performance suffers from two limitations: handling extreme lighting scenes with both highlight and low-light areas and robustness to camera motion blur, which are caused by the inherent information loss in the blurred and compressed LDR images with both underexposed and overexposed regions.
Some recent image-based NeRF and 3DGS related works utilize multi-exposure LDR images~\cite{Evhdr-NeRF, hdr-gs, gausshdr} to compensate for the limited dynamic range and model the camera motion~\cite{deblurenerf, bad-nerf, bad-gs} to mitigate the influence of motion blur,
while they are still susceptible to the requirement of cumbersome multi-exposure capture and severe motion blur under long-exposure conditions.

Event cameras provide a promising sensing modality by asynchronously recording brightness changes in the log domain, thereby preserving high temporal resolution and dynamic range information that is typically lost in conventional frame-based RGB cameras~\cite{event}.
As shown in the upper half of \cref{fig:1}, event data can effectively record spatial-temporal differential data even in blurred, overexposed, and underexposed areas of an LDR image.
Motivated by this advantage, recent approaches have explored using event and RGB image (ERGB) data for the deblurring NVS task~\cite{e2nerf, e3nerf, evdeblurnerf, ebad-nerf, e2gs}, while their performance is limited by a lack of grounding in the sensor-physics mismatch between the recorded RGB image and the physical scene brightness.
EvHDR-NeRF~\cite{Evhdr-NeRF} first incorporates events into the single-exposure HDR and deblurring NeRF reconstruction task.
However, its framework largely follows HDR-NeRF~\cite{hdrnerf} by applying the same tone mapping strategy directly to 3D points and extending it to the event-based setting without considering the event generation character.

We argue that achieving accurate and physically consistent 3D representation learning in the ERGB setting requires more than directly extending existing image-based frameworks.
In particular, it is essential to explicitly model how scene radiance is transformed into actual sensor measurements through device-specific processes. For RGB cameras, this involves temporal integration and non-linear camera response functions (CRF); for event cameras, this includes contrast thresholding, latency effects, and photometric quantization. Without accounting for these physical transformations, the rendered outputs of the learned model may not align with the supervision signals derived from input images and events, weakening the learning of actual scene geometry and appearance as shown in the lower half of \cref{fig:1}.
Therefore, a sensor-physics grounded modeling of both RGB and event sensing pipelines is critical for fully exploiting the complementary information in ERGB data and achieving robust deblurring HDR NVS performance.

In this paper, we present a novel framework that exploits \textbf{S}ingle-\textbf{E}xposure blurry images and corresponding \textbf{E}vents to enable sharp HDR NeRF reconstruction (\textbf{See-NeRF}).
Specifically, built upon the physical imaging principles of RGB and event sensors, we adopt NeRF to focus on learning the actual radiance of the scene in the HDR domain and introduce two differentiable CRF models: a 2D pixel-wise RGB CRF model that accounts for exposure integration and camera response function in image generation, and a novel event CRF model that accounts for latency and photometric deviations in event generation.
These two modules align the predicted radiance with the sensor measurements and are jointly optimized with the NeRF network under the supervision of input images and events.
By modeling the radiance-to-measurement process explicitly, See-NeRF can accurately reconstruct physically consistent, sharp HDR 3D representations, achieving state-of-the-art deblurring HDR NVS results on both public and collected datasets.
Our contributions are summarized as follows:

1) We propose See-NeRF, a sensor-physics grounded framework that uses events to compensate for the scene dynamics to learn a sharp HDR 3D representation from single-exposure blurry LDR images in extreme lighting scenes.

2) We introduce a physically grounded RGB CRF model and a latency-aware photometrically calibrated event CRF model to bridge the discrepancy between physical radiance values and sensor measurements.

3) We collect both synthetic and real datasets for training and testing.
The experiment results show that our method achieves the best performance on both ERGB-based  HDR and deblurring novel view synthesis tasks.


\section{Related work}
\label{sec:2}

\subsection{HDR and deblurring novel view synthesis}
\label{sec:2.1}
NeRF~\cite{nerf} extensively promotes the development of the novel view synthesis (NVS) task and 3D implicit representation learning.
3DGS~\cite{3DGS} further improve the speed of 3D reconstruction and view rendering.
However, NeRF and 3DGS can only learn a sharp LDR scene representation with sharp LDR images.
For the HDR NVS task, RawNeRF~\cite{rawnerf} and RawGS~\cite{rawgs} aim to reconstruct an HDR 3D representation from noisy HDR raw images.
HDR-NeRF~\cite{hdrnerf} and HDR-GS~\cite{hdr-gs} introduce a 3D points tone mapper to transform the HDR 3D representation into the LDR 3D representation, optimized with multi-exposure LDR images.
HDR-HexPlane~\cite{hdr-hexPplane} further extends HDR-NeRF to dynamic scenes.
GaussHDR~\cite{gausshdr} accepts additional context features as input to improve HDR NVS.
Gaussian in the dark~\cite{gaussian-in-the-dark} uses a CNN-based feature map to enhance the tone mapper learning.
For the Deblurring NVS task, Deblur-NeRF~\cite{deblurenerf} proposes a deformable sparse kernel module to model the blur kernels.
BAD-NeRF~\cite{bad-nerf} and BAD-GS~\cite{bad-gs} transform the camera poses into SE(3) space and interpolate the poses to recover the camera motion trajectory.
DP-NeRF~\cite{dpnerf} imposes 3D consistency and refines the color with the relationship between depth and blur.
Although these image-based methods have achieved promising HDR and deblurring NVS performance, they still fundamentally suffer from the inherent information loss of the motion-blurred LDR images.

\subsection{ERGB-based novel view synthesis}
\label{sec:2.2}
Event cameras are bio-inspired sensors detecting brightness changes at individual pixels with high temporal resolution~\cite{event}.
This scene radiance measurement paradigm has higher dynamic range and temporal resolution compared to traditional frame-based RGB cameras, attracting a lot of research in the fields of event-based image deblurring~\cite{evdeblur1,evdeblur2,evdeblur3,evdeblur4,evdeblur5}, HDR image reconstruction~\cite{evhdr0, evhdr1, evhdr2, evhdr3, evhdr4, evhdr5, evhdr6}, and high-speed HDR video generation~\cite{evhdr-video0, evhdr-video1, evhdr-video2, evhdr-video3}.
Some recent works explore the event-based NVS task~\cite{enerf1,eventnerf,renerf,deblurenerf,aenerf,Elite-EvGS,event-3dgs,event-3dgs,eventsplat}, while other works integrate both image and event (ERGB) data into the deblurring NVS task, such as NeRF-based works~\cite{e2nerf, e3nerf,evdeblurnerf,ebad-nerf, benerf, lsenerf} and 3DGS-based works~\cite{e2gs,ev3dgs,evagaussians,dietgs}.
Addressing the limitations of blurry LDR inputs, EvHDR-NeRF~\cite{Evhdr-NeRF} first jointly exploits single-exposure images and events to reconstruct HDR NeRF.
However, these previous ERGB-based NVS methods do not fully consider the gap between sensor output and physical radiance, leading to limited HDR and deblurring NVS performance.

\section{Background}
\label{sec:3}
In this section, we leverage the principles of event and LDR image generation to explain why events can enhance HDR deblurring NVS with single-exposure blurry LDR images.

\subsection{From scene radiance to LDR images}
\label{sec:3.1}
In an extreme lighting scene with both highlight and low-light areas, a typical RGB camera sensor records the physical scene radiance through a frame-based integral model, generating a raw image with high dynamic range pixel values.
To fit human eye perceptual characteristics and reduce storage costs, the raw image is processed by the image signal processor (ISP), applying non-linear ISO gain, RGB demosaicing, tone mapping, gamma compression, white-balance, and quantization to yield an 8-bit LDR image, which is always modeled by a camera response function (CRF) $f$ in practice~\cite{isp}:
\begin{equation}
    \mathcal{I}_{\text{LDR}} = f(\int_{t_{\text{start}}}^{t_{\text{end}}} L(t) dt),
    \label{eq:1}
\end{equation}
where $L(t)$ is the raw scene radiance hitting the sensor at time $t$.
$t_{\text{start}}$ and $t_{\text{end}}$ are the start and end time of exposure.
$\mathcal{I}_{\text{LDR}}$ is the final color value that we can obtain from the LDR image.
This integral model and non-linear CRF $f$ inherently obscure the temporal dynamics of $L(t)$ and compress the critical HDR spatial radiance differential information in highlight and low-light areas, ineluctably leading to artifacts like motion blur, overexposure (pixel values saturation at 255), and underexposure (pixel values clipping at 0) in the LDR images, respectively.

\begin{figure}[t]
  \centering
  \includegraphics[width=1\linewidth]{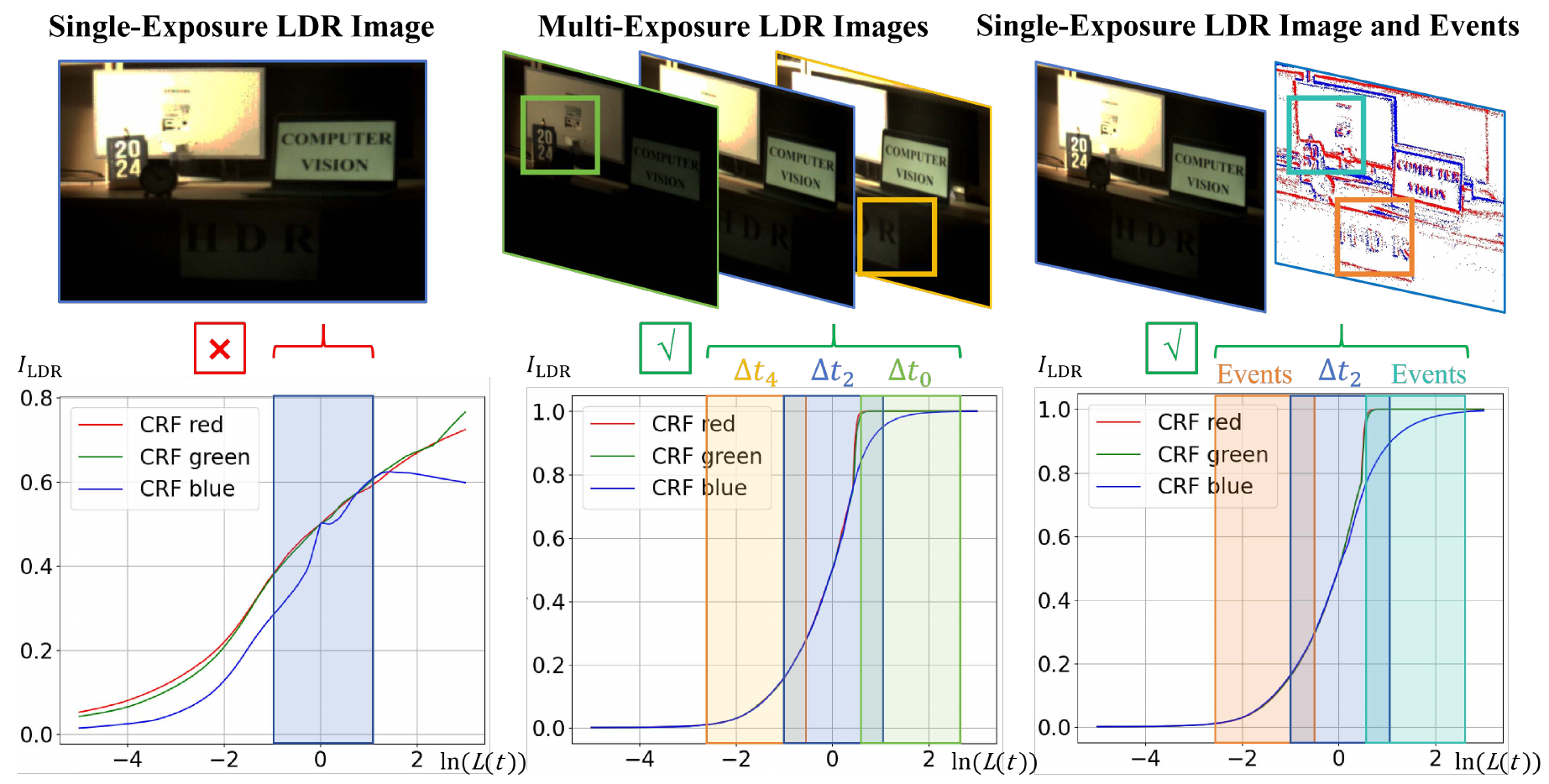}
   \caption{HDR reconstruction from Events. The first row shows the input data. The second row shows the estimated CRF curves and the dynamic range schematic. For a single-exposure LDR image with exposure time $\Delta t_2$, its 8-bit pixel values 0-255 (normalized to 0-1 as $\mathcal{I}_{\text{LDR}}$ in the figure) represent a limited range of scene radiance $\ln(L(t))$ (width of the blue box). Traditional methods fuse multi-exposure LDR images with exposure times $\Delta t_0, \Delta t_4$ to compensate for the limited scene brightness representation (width of the yellow and green boxes). We leverage the spatial differential in events to infer the scene radiance of highlight and low-light regions from the single-exposure LDR image ($\Delta t_2$), which expands the representation of scene radiance (width of the orange and cyan boxes), enabling more accurate CRF curves estimation.}
   \label{fig:2}
\end{figure}

\subsection{From scene dynamics to events}
\label{sec:3.2}
Event cameras asynchronously detect the change of the scene radiance hitting the sensor $L(t)$ in the log domain and trigger an event $e(x,y,p,\tau)$ whenever the change reaches the threshold $\Theta$.
$(x,y)$ represents the pixel coordinates and $\tau$ is the trigger time.
The change direction is encoded in $p\in\{1,-1\}$.
This spike model can preserve temporal radiance dynamics and spatial radiance gradient distribution:
\begin{equation}
    \begin{aligned}
    \text{Temporal:} \partial_t \log L(x,y,t),\\
    \text{Spatial:} \nabla_{xy} \log L(x,y,t),
    \end{aligned}
    \label{eq:2}
\end{equation}
which precisely compensates for the temporal and spatial dynamics loss of blurry compressed LDR images (\cref{sec:3.1}).

\subsection{Events for deblurring and HDR NVS}
According to \cref{sec:3.1}, the core problem of recovering the sharp HDR scene radiance from blurry LDR images is formulated as the joint estimation of the latent sharp luminance $L(t)$ and the CRF $f$.
The temporal radiance dynamics of events in \cref{eq:2} can be used to estimate the latent sharp $L(t)$ along the time dimension from blurry $\mathcal{I}_{\text{LDR}}$ as proved in ERGB-based deblurring works~\cite{evdeblur1, e2nerf}.
The spatial dynamics of events in \cref{eq:2} can provide extra radiance differential information for the highlight and low-light areas and expand the dynamic range of the single-exposure LDR image $\mathcal{I}_{\text{LDR}}$, as shown in \cref{fig:2}, with which the network in \cref{sec:4.2} can obtain a wider range of $\ln(L(t))$ as input and infer a CRF curve closer to the ground truth.


\begin{figure*}[t]
  \centering
  \includegraphics[width=1\linewidth]{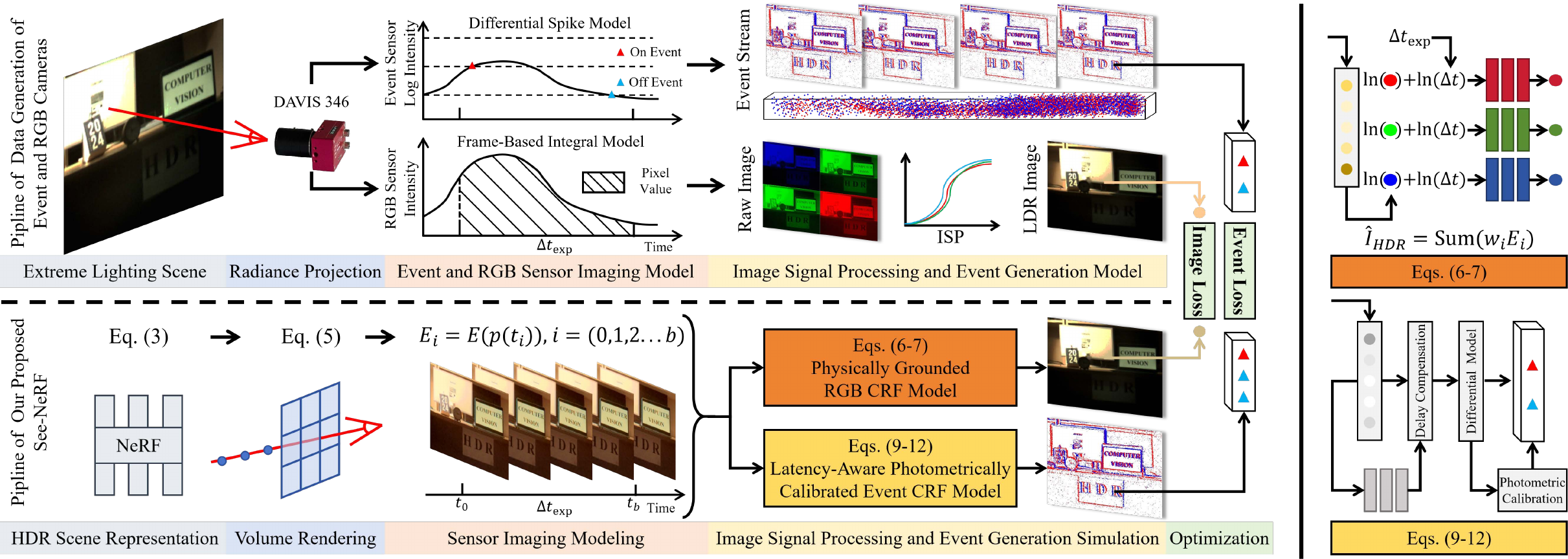}
    \caption{The data generation process of event and RGB cameras (upper parts) and the pipeline of See-NeRF (lower parts) are shown in the figure. We use NeRF to represent the actual radiance of an extreme lighting scene in the HDR domain. The volume rendering simulates the raw scene radiance rays hitting the sensor and obtaining the raw sensor values. The predicted LDR images and events are generated with our proposed sensor-physics grounded RGB and event CRF models (details are shown in the right part of the figure).
   The image loss and event loss are employed to jointly supervise the optimization of the two CRF models and the NeRF network.}
   \label{fig:3}
   \vspace{-1em}
\end{figure*}

\section{Method}
\label{sec:4}
In this section, we introduce the See-NeRF and theoretically demonstrate how it leverages sensor-physics priors to achieve high-quality deblurring HDR NVS with single-exposure blurry LDR images and corresponding events
 
\subsection{HDR scene representation}
\label{sec:4.1}
Firstly, we use the NeRF network $F_{\theta}$ with parameters $\theta$ to represent the actual radiance $\mathbf{e}$ and density $\sigma$ of a 3D point $\mathbf{o}$ in an extreme lighting scene, rather than the camera-processed LDR radiance used in the original NeRF~\cite{nerf}:
\begin{equation}
    (\textbf{e}, \sigma)=F_{\theta}(\gamma_{o}(\textbf{o}), \gamma_{d}(\textbf{d})),
    \label{eq:3}
\end{equation}
where $\textbf{d}$ denotes the observation direction and $\gamma (\cdot)$ is used to transform the input into a higher dimension:
\begin{equation}
    \gamma_{M}(x) = \{\sin(2^{m}\pi{x}), \cos(2^{m}\pi{x})\}_{m=0}^{M}.
    \label{eq:4}
\end{equation}
Then we use volume rendering to simulate the process of HDR scene rays $\mathbf{o}+\mathbf{d}$ hitting the imaging sensor at pixel $(x,y)$ corresponding to the camera pose $p(t)$ and obtain the latent sharp HDR radiance at time $t$:
\begin{equation} 
    \begin{aligned}
        E(x,y,p(t)) = \sum_{i=1}^{N}T_{i}(1-\exp(-{\sigma}_i{\delta}_i))\mathbf{e}_i,\\
        \text{where} \quad T(i) = \exp(-\sum_{j=1}^{i-1}\sigma_{j}\delta_{j}).
    \end{aligned}
\label{eq:5}
\end{equation}
$\delta_{i}$ is the distance between adjacent sampled 3D points.

\subsection{Physically grounded RGB CRF model}
\label{sec:4.2}
To simulate the integral model of the RGB camera, we discretely sample $b+1$ time points $\{{t_i}\}_{i=0}^{b}$ within the exposure time $t_{\text{start}}$ to $t_{\text{end}}$ as in~\cite{e3nerf}, and the predicted raw pixel value of the RGB imaging sensor during exposure is:
\begin{equation} 
    \begin{aligned}
        \hat{\mathcal{I}}_{\text{HDR}}(x,y) = \sum_{i=0}^{b}w_{i}E_{i}(x,y,p(t_i)),
    \end{aligned}
\label{eq:6}
\end{equation}
where $w_i$ is weight parameters calculated by time points $\{t_{i}\}_{i=0}
^{b}$~\cite{e3nerf}.
Since $p(t)$ is constantly changing during exposure in the handheld situation, we use the event-guided COLMAP~\cite{colmap1} strategy in~\cite{e2nerf,e3nerf} for the pose estimation.
To bridge the gap between the raw sensor value and the LDR image calculated by the CRF $f$ in \cref{eq:1}, we used three MLPs $f_{\text{crf}}$ to fit this function in the log domain for three color channels separately and obtain the estimated LDR pixels:
\begin{equation} 
    \begin{aligned}
        \hat{\mathcal{I}}_{\text{LDR}}(x,y) = f_{\text{crf}}(\text{ln}(\hat{\mathcal{I}}_{\text{HDR}}(x,y) \Delta{t}_{\text{exp}})),
    \end{aligned}
\label{eq:7}
\end{equation}
where $\Delta{t}_{\text{exp}}=t_{\text{end}}-t_{\text{start}}$ is the exposure time.

\begin{figure}[t]
  \centering
  \includegraphics[width=1\linewidth]{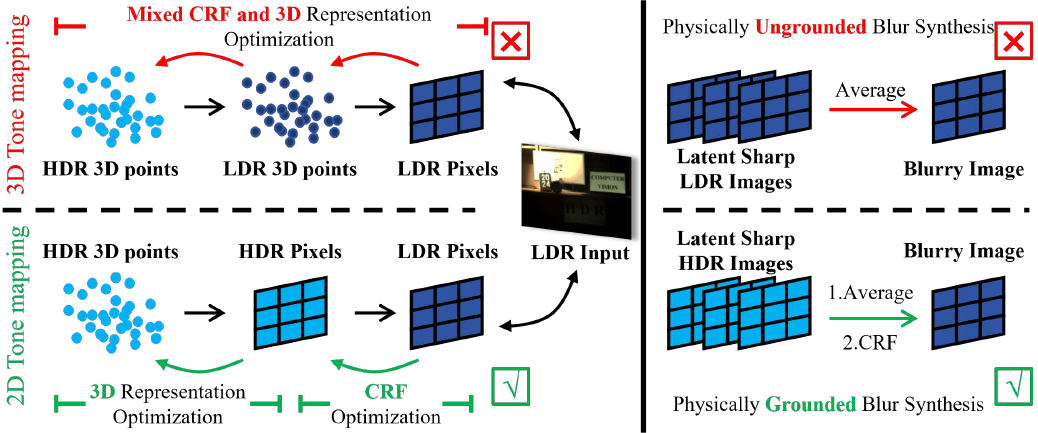}
   \caption{Effectiveness of physically grounded RGB CRF model.}
   \label{fig:4}
   \vspace{-1em}
\end{figure}

As shown in \cref{fig:4}, unlike HDR-NeRF applying the CRF function directly on the radiance of 3D points $\mathbf{e}$ in \cref{eq:3}, we employ CRF after volume rendering at the pixel-level for the raw pixel values $\hat{\mathcal{I}}_{\text{HDR}}$, aligning with the imaging process in the physical world (\cref{fig:3}), which allows $f_{\text{crf}}$ to focus on non-linear color tone mapping learning without being affected by the linear volume rendering, while NeRF $F_{\theta}$ focuses on learning 3D density $\sigma$ and raw radiance $\mathbf{e}$ without being affected by $f_{\text{crf}}$.
Besides, our blur synthesis on the HDR pixel values (\cref{eq:6}) is more physically grounded.
Therefore, our proposed RGB CRF model can significantly improve the HDR and deblurring NVS results as demonstrated by the experiment in \cref{sec:5.4}.

\subsection{Latency-aware and photometrically calibrated event CRF model}
\label{sec:4.3}
Usually, we can estimate the generated event number during $t_1$ and $t_2$ at pixel $(x,y)$ with an ideal model:
\begin{equation}
    B(t_1,t_2,x,y) = \text{floor}(\frac{\ln(L(t_2)) - \ln(L(t_1))}{\Theta}),
    \label{eq:8}
\end{equation}
where $\text{floor}(\cdot)$ means taking the integer between the input and zero that is closest to the input.
However, the actual event generation may deviate from this ideal model with a significant delay in low-light regions~\cite{v2e}, and due to the fixed contrast threshold $\Theta$ in event generation, the minimum detectable brightness change is lower-bounded, leading to quantifiable errors in radiometric estimation.
To bridge the gap between actual radiance dynamics and actual event generation, we propose a novel event CRF model comprising three core components: Bayer pattern adaptation, temporal delay compensation, and photometric quantity calibration.

\myPara{Bayer pattern adaptation:}
Like in~\cite{evdeblurnerf, Evhdr-NeRF}, we input the $b+1$ scene radiance values $\{E(x,y,p(t_i))\}_{i=0}^{b}$ with three color channels into an RGGB Bayer pattern to obtain the pixel radiance values of the event sensor $\{E_{\text{ev}}(x,y, p(t_i))\}_{i=0}^{b}$.

\myPara{Temporal delay compensation:}
Since event delay is mainly determined by the pixel radiance values~\cite{v2e}, we calculate the delay coefficient with a MLP-based function $f_{\text{ev}}$:
\begin{equation} 
    \begin{aligned}
       \epsilon_{i} = f_{\text{ev}}(E_{\text{ev}}(x,y,p(t_{i}))).
    \end{aligned}
\label{eq:9}
\end{equation}
Then, we use a second-order low-pass filter to synthesize the event latency with the estimated coefficient $\epsilon_{i}$:
\begin{equation} 
    \begin{aligned}
        L^{i}_{\text{lp}} &= (1-\epsilon_{i})L^{i-1}_{\text{lp}}+\epsilon_{i} E_{\text{ev}}(p(t_i)).
    \end{aligned}
\label{eq:10}
\end{equation}
We omit $(x,y)$ in \cref{eq:10} because the above operation is the same for all pixels.
We set $L^0_{\text{lp}}=E_{\text{ev}}(p(t_0))$ and substitute $\{ L^i_{\text{lp}}\}^b_{i=0}$ into the event generation model (\cref{eq:8}) to calculate the estimated numbers of events:
\begin{equation}
    \{\hat{B}'(t_{i}, t_{i+1})\}_{i=0}^{b-1} = \text{floor}(\frac{\ln(L^{i+1}_{\text{lp}}) - \ln(L^{i}_{\text{lp}})}{\Theta}).
    \label{eq:11}
\end{equation}

\myPara{Photometric quantity calibration:}
Since event acquisition exhibits uncertainty when brightness changes are below the threshold or our sampled time points $t_i$ may not precisely align with the timestamp of the last triggered event at this pixel, we pre-estimate an offset from the input events by using $h(\cdot)$, and then subtract this offset for the final estimated number of events to compensate this discrepancy:

\begin{equation}
    \{\hat{B}(t_{i}, t_{i+1})\}_{i=0}^{b-1} = \hat{B}'(t_{i}, t_{i+1}) - h({{B}}(t_{i}, t_{i+1})),
    \label{eq:12}
\end{equation}
$h(\cdot)$ is explained in detail in the supplementary materials.

\subsection{Loss functions and optimization}
\label{sec:4.4}
We use the LDR image loss and event loss to jointly supervise the optimization of the HDR scene radiance representation (\cref{eq:3}) and the two CRF models:
\begin{equation}
    \mathcal{L}_{\text{ldr}} = \sum_{(x,y)\in\mathcal{X}}||\hat{\mathcal{I}}_{\text{LDR}}(x,y)-\mathcal{I}_{\text{LDR}}(x,y)||^2_2,
    \label{eq:13}
\end{equation}
\begin{equation}
    \mathcal{L}_{\text{evs}} = \sum_{(x,y)\in\mathcal{X}}\sum_{i=0}^{b-1}||\hat{B}(t_{i}, t_{i+1})-B(t_{i}, t_{i+1})||^2_2,
    \label{eq:14}
\end{equation}
where $\mathcal{X}$ denotes all pixels of the camera sensor.
Then the final loss is defined as:
\begin{equation}
    \mathcal{L} = \lambda\mathcal{L}_{\text{evs}} +\mathcal{L}_{\text{ldr}}.
    \label{eq:15}
\end{equation}
We take $\lambda=0.005$ and optimize a coarse and a fine network simultaneously as in~\cite{e3nerf}.
During testing, we can obtain the novel view sharp HDR images with \cref{eq:6} and LDR images with novel exposure with \cref{eq:7}.
We set $b=4$ as a trade-off between performance and training time as in~\cite{e2nerf}.
The implementation details and more evaluation on $b$ and $\lambda$ are provided in Sec. C and  Sec. D of the supplementary.

\begin{table*}[t]
    \centering
    \caption{Quantitative results on our synthetic and real dataset for the HDR and novel exposure NVS tasks. The results are the average of all scenes in the dataset. HDR-NeRF\textsuperscript{ref}: training HDR-NeRF~\cite{hdrnerf} with multi-exposure input. HDR-NeRF+: training HDR-NeRF with sharp HDR images obtained by the cascaded ERGB-based image deblurring method EDI~\cite{evdeblur1} and HDR reconstruction method HDRev~\cite{evhdr-video2}.}
    \label{tab:1}
    \scriptsize
    \setlength{\tabcolsep}{0.35mm}
        \begin{tabular}{c|l|cc|ccc|ccc|ccc|ccc}
        \toprule
        & & & & \multicolumn{6}{c}{Our Synthetic Dataset} & \multicolumn{6}{|c}{Our Real Dataset} \\
        & & \multicolumn{2}{c|}{Input} 
        & \multicolumn{3}{c}{HDR} 
        & \multicolumn{3}{|c}{Novel Exposure ($\Delta t_{1}$, $\Delta t_{3}$)} 
        & \multicolumn{3}{|c}{HDR} 
        & \multicolumn{3}{|c}{Novel Exposure ($\Delta  t_{1}$, $\Delta t_{3}$)}  \\
        \midrule
        & Methods & Image & Event & PSNR\textuparrow & SSIM\textuparrow & LPIPS\textdownarrow & PSNR\textuparrow & SSIM\textuparrow & LPIPS\textdownarrow & PSNR\textuparrow & SSIM\textuparrow & LPIPS\textdownarrow & PSNR\textuparrow & SSIM\textuparrow & LPIPS\textdownarrow \\
        
        \midrule
        Reference & HDR-NeRF\textsuperscript{ref} & Sharp ($\Delta t_0$, $\Delta t_2$, $\Delta t_4$) & - & 26.91 & .8906 & .1705 & 29.22 & .9247 & .1265 & 26.26 & .8716 & .1576 & 33.44 & .9618 & .0950 \\

        \midrule
        \multirow{4}{*}{\makecell[c]{RGB-Based \\ HDR NVS}}
        & HDR-NeRF~\cite{hdrnerf}        & Sharp ($\Delta t_2$) & - & 17.02 & .7384 & .3450 & 24.61 & .9178 & .1349 & 10.87 & .5473 & .4229 & 22.18 & .9013 & .1230 \\
        & HDR-GS~\cite{hdr-gs}          & Sharp ($\Delta t_2$) & - & 13.54 & .6993 & .5096 & 15.13 & .7428 & .3533 & 14.18 & .6944 & .4700 & 19.23 & .8407 & .1980 \\
        
        & Gaussian-DK~\cite{gaussian-in-the-dark} & Sharp ($\Delta t_2$) & -  & 12.74 & .6826 & .3206 & 25.87 & .9197 & .1792 & 14.33 & .7113 & .4055 & 26.41 & .8966 & .1184 \\
        & GaussHDR~\cite{gausshdr}          & Sharp ($\Delta t_2$) & -  & 12.96 & .7805 & .3015 & 14.69 & .7565 & .2154 & 20.39 & .8669 & .1723 & 26.77 & .9361 & .1771 \\
        
        \midrule
        \multirow{3}{*}{\makecell[c]{ERGB-Based \\ Deblurring \\ HDR NVS}}
        & HDR-NeRF+       & Blurry ($\Delta t_2$) & \Checkmark & 16.08 & .6436 & .3696 & 17.63 & .7178 & .2786 & 18.61 & .8092 & .2872 & 15.30 & .5936 & .3444 \\
        & EvHDR-NeRF~\cite{Evhdr-NeRF}      & Blurry ($\Delta t_2$) & \Checkmark & 21.73 & .8192 & .3446 & 24.37 & .8712 & .2848 & 19.00 & .6962 & .2612 & 23.76 & .8830 & .1650 \\
        & See-NeRF        & Blurry ($\Delta t_2$) & \Checkmark & \textbf{24.13} & \textbf{.8927} & \textbf{.1916} & \textbf{27.57} & \textbf{.9270} & \textbf{.1328} & \textbf{26.49} & \textbf{.8953} & \textbf{.1638} & \textbf{29.11} & \textbf{.9364} & \textbf{.1060} \\
        \bottomrule
    \end{tabular}
    \vspace{-1em}
\end{table*}

\begin{figure*}[t]
    \centering
    \includegraphics[width=1\linewidth]{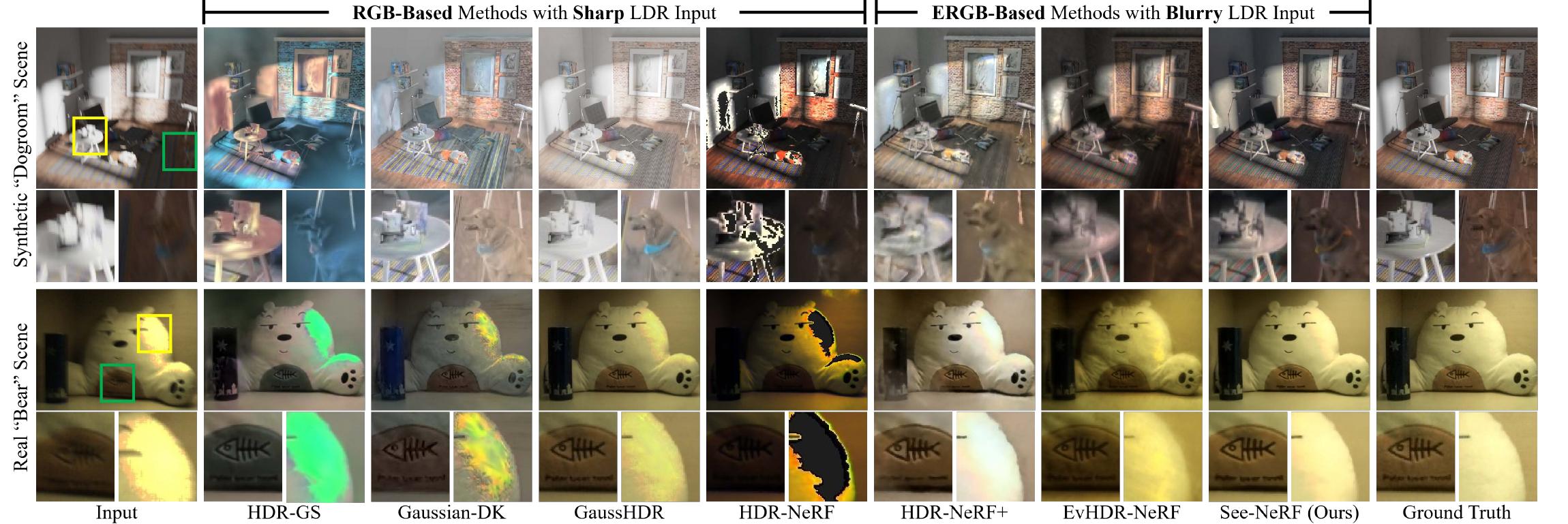}
    \caption{Qualitative results on the synthetic ``dogroom'' scene and real-world ``bear'' scene of our datasets for the HDR novel view synthesis task. Our See-NeRF achieves the best results on both under-exposure and over-exposure regions.}
    \label{fig:5}
    \vspace{-1em}
\end{figure*}

\begin{figure*}[t]
    \centering
    \includegraphics[width=1\linewidth]{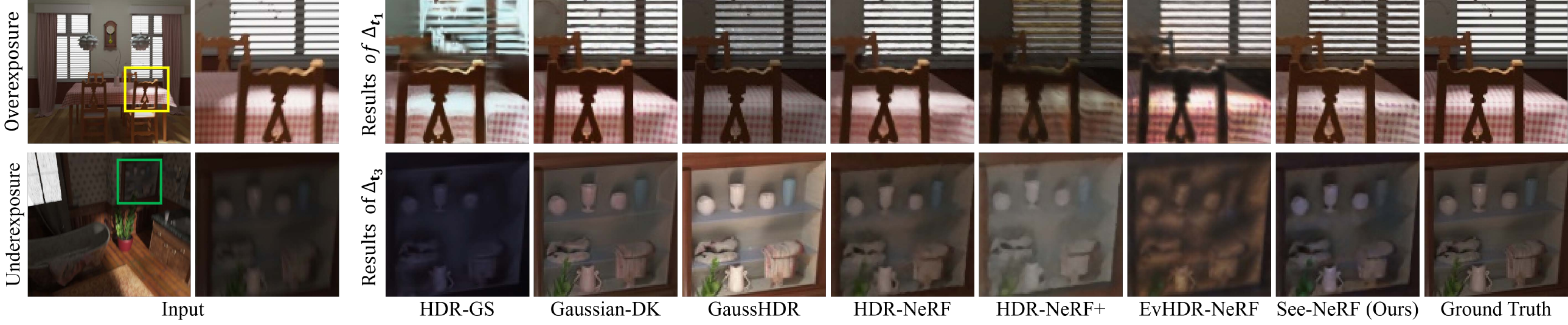}
    \caption{Qualitative results on the synthetic ``diningroom'' (upper) and ``bathroom'' (lower) scenes for novel exposure synthesis task.}
    \label{fig:6}
    \vspace{-1em}
\end{figure*}


\section{Experiments}
\label{sec:5}
Since EvHDR-NeRF~\cite{Evhdr-NeRF} has not released its datasets, we construct both synthetic and real datasets for the deblurring HDR NVS task from single-exposure blurry LDR images and corresponding events.
We also adopt the public Real-World-Challenge dataset~\cite{e3nerf} with challenging blurry ERGB input and GT sharp test images to evaluate the deblurring effect of See-NeRF.
Due to space limitations, we only briefly describe the collected dataset here and provide further dataset generation and experimental details in the Secs. B, C, and D of the supplementary materials.

\subsection{Experiment settings}
\label{sec:5.1}

\myPara{Synthetic dataset:}
The synthetic dataset consists of 8 HDR ``Blender'' scenes from HDR-NeRF~\cite{hdrnerf}.
Each scene contains 18 training and 17 test views.
We use Blender~\cite{Blender} and v2e~\cite{v2e} to generate blurry images with exposure time $\Delta t_2$ and corresponding events as in~\cite{e2nerf} for training views.
We also provide sharp LDR images with exposure times $\Delta t_0, \Delta t_2, \Delta t_4$ for the RGB-based HDR NVS methods without deblurring effects.
For test views, we provide GT sharp HDR images and LDR images with $\Delta t_1, \Delta t_3$ for HDR and novel exposure evaluation as in~\cite{hdrnerf}.

\myPara{Real dataset:}
The real dataset includes 5 scenes captured under extreme lighting conditions using the DAVIS 346 event camera~\cite{davis346}.
Each scene has 16 training and 12 test views.
For training views, we use a handheld DAVIS 346 to capture the blurry LDR images with exposure time $\Delta t_2$ along with the corresponding event data.
We also captured sharp LDR images with a tripod with exposure times $\Delta t_0, \Delta t_2, \Delta t_4$.
For the test views, we use a tripod to capture GT sharp LDR images with exposure times $\Delta t_1$, $\Delta t_3$.
We also capture the GT sharp LDR images with $\Delta t_0$, $\Delta t_2$, $\Delta t_4$, and use the classic multi-exposure HDR reconstruction algorithm DeBevec~\cite{hdr01} to generate the GT sharp HDR image with the 5 sharp LDR images with different exposures for each test view as in RawHDR~\cite{single-raw-hdr}.

\myPara{Metrics:}
We use PSNR, SSIM, and LPIPS~\cite{lpips} to quantitatively evaluate the deblurring and novel exposure NVS results.
For the HDR NVS task, we tone map~\cite{Photomatix} both the generated and GT HDR images into the LDR domain for quantitative and qualitative evaluation as in~\cite{hdr-gs,hdrnerf,single-raw-hdr}.

\subsection{HDR and novel exposure NVS experiment}
\label{sec:5.2}
We choose the RGB-based HDR NVS works~\cite{hdrnerf, hdr-gs, gausshdr, gaussian-in-the-dark} for comparison with sharp single-exposure images as input, considering they cannot handle the blurry images.
We also use multi-exposure images as input for HDR-NeRF as a reference (HDR-NeRF\textsuperscript{ref}).
For the ERGB-based methods, we use EDI~\cite{evdeblur1} and HDRev~\cite{evhdr-video2} to deblur the blurry LDR images and generate sharp HDR images with event data, which are input into the HDR-NeRF for training (HDR-NeRF+).
We also compare with the ERGB-based deblurring HDR NeRF method, EvHDR-NeRF.

\myPara{Quantitative results:}
As in \cref{tab:1}, our See-NeRF significantly outperforms the compared methods on both synthetic and real datasets.
The HDR reconstruction results on the real dataset are even better than HDR-NeRF\textsuperscript{ref}.
In comparison, RGB-based HDR NVS methods exhibit constrained performance owing to their dependence on multi-exposure input.
Cascaded method HDR-NeRF+ achieves superior results compared to RGB-based methods but underperforms EvHDR-NeRF due to its decoupled pipeline that separately handles 3D reconstruction and HDR sharp image generation, constrained by the inherent bottlenecks of EDI and HDRev.
EvHDR-NeRF realizes the second-best overall results, but is significantly worse than our See-NeRF.

\myPara{Qualitative results:}
\cref{fig:5} displays the qualitative HDR NVS results, and our See-NeRF achieves the best performance.
The results of EvHDR-NeRF exhibit slight color deviation and blur artifacts,
which stem from inaccurate CRF estimation and unmodeled radiance integration of the RGB sensor.
HDR-NeRF+ generates results with elevated luminance but significant color distortion, a limitation primarily attributable to suboptimal radiance estimation of HDRev.
Lack of multi-exposure input, RGB-based methods have serious color casts, especially in the overexposed parts of the real data.
For the novel exposure NVS task, See-NeRF perfectly recovers both overexposed and underexposed areas, while other methods are affected by color and brightness mismatch as shown in \cref{fig:6}.

\begin{table}[t]
    \centering
    \caption{Results on the Real-World-Challenge dataset of the deblurring NVS task. The results are the average of all five scenes.}
    \scriptsize
    \setlength{\tabcolsep}{1.5mm}
        \begin{tabular}{c|l|ccc}
        \toprule
        &Methods & PSNR\textuparrow & SSIM\textuparrow & LPIPS\textdownarrow \\
        \midrule
        \multirow{2}{*}{\makecell[c]{RGB-Based \\ Deblurring NVS}}
        & BAD-NeRF~\cite{bad-nerf} & 25.63 & .8575 & .4400 \\
        & DP-NeRF~\cite{dpnerf} & 28.85 & .9226 & .3015 \\
        \midrule
        \multirow{5}{*}{\makecell[c]{ERGB-Based \\ Deblurring NVS}}
        
        & EBAD-NeRF~\cite{ebad-nerf}             & 27.74 & .9010 & .3349 \\
        & Ev-DeblurNeRF~\cite{evdeblurnerf}      & 27.83 & .9132 & .3157 \\
        & E2GS~\cite{e2gs}                       & 28.04 & .9211 & .2538 \\
        & E\textsuperscript{2}NeRF~\cite{e2nerf} & 29.92 & .9346 & .2356 \\
        & E\textsuperscript{3}NeRF~\cite{e3nerf} & 31.40 & .9464 & .2000  \\
        \midrule

        \multirow{2}{*}{\makecell[c]{ERGB-Based \\ Deblurring HDR NVS}}
        & EvHDR-NeRF~\cite{Evhdr-NeRF}           & 27.19 & .8960 & .3731 \\
        & See-NeRF (Ours)                        & \textbf{32.70} & \textbf{.9564} & \textbf{.1574}\\
        \bottomrule
        \end{tabular}
    \label{tab:2}
    \vspace{-1em}
\end{table}

\begin{figure}[t]
    \centering
    \includegraphics[width=1\linewidth]{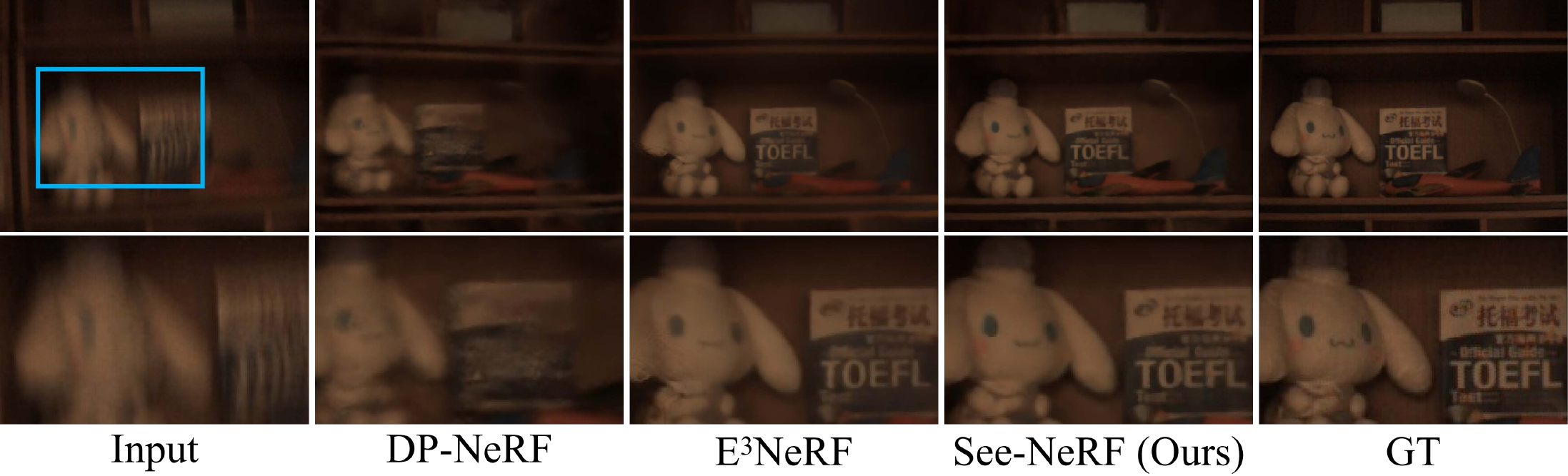}
    \caption{Qualitative results on the ``shelf'' scene of the Real-World-Challenge data for the deblurring NVS task.}
    \label{fig:7}
    \vspace{-1em}
    \end{figure}

\subsection{Deblurring NVS experiment}
\label{sec:5.3}

In this experiment, we compared our method with recent RGB-based and ERGB-based deblurring NVS works, and the quantitative results are shown in \cref {tab:2}.
Our See-NeRF achieves significant performance improvements, especially on the LPIPS, proving that our proposed event and RGB CRF models effectively bridge the sensor-physics discrepancy, strengthening the use of non-ideal data to reconstruct the sharp 3D scene.
\cref{fig:7} shows that See-NeRF successfully recovers the blue eyes and two pink cheek dots of the rabbit, while the best ERGB-based deblurring NVS method E\textsuperscript{3}NeRF misses these subtle details.
The results of the best RGB-based deblurring NVS method, DP-NeRF, are inferior due to the lack of event enhancement.

\begin{table*}[t]
    \centering
    \caption{Quantitative results of our ablation study. The results are the averages of all scenes.}
    \scriptsize
    \setlength{\tabcolsep}{1.2mm}
        \begin{tabular}{l|ccc|ccc|ccc|ccc}
        \toprule
        &
        & 
        & 
        & \multicolumn{3}{c|}{Our Synthesis Data (HDR)} 
        & \multicolumn{3}{|c}{Our Real Data (HDR)}
        & \multicolumn{3}{|c}{Real-World-Challenge Data} 
         \\
        & Event Input & Event CRF Model & RGB CRF model & PSNR\textuparrow & SSIM\textuparrow & LPIPS\textdownarrow & PSNR\textuparrow & SSIM\textuparrow & LPIPS\textdownarrow & PSNR\textuparrow & SSIM\textuparrow & LPIPS\textdownarrow  \\
        
        \midrule

        See-NeRF\textsuperscript{Base}        & -             & -         & Naive~\cite{nerf,e2nerf,e3nerf}  & - & - & - & - & - & - & 28.79 & .9201 & .2939 \\
        
        See-NeRF\textsuperscript{NoEv}        & -             & -         & ours  & 22.19 & .8361 & .2600 & 19.84 & .7881 & .2502 & 30.22 & .9350 & .2425 \\

        \midrule
        See-NeRF\textsuperscript{NoEM}         & \Checkmark   & Naive~\cite{e2nerf,e3nerf,Evhdr-NeRF}  & ours  & 22.23 & .8682 & .2254 & 25.87 & .8867 & .1701 & 32.47 & .9547 & .1686\\
        See-NeRF\textsuperscript{3D}         & \Checkmark     & ours    & 3D Points CRF~\cite{hdrnerf,hdr-gs,Evhdr-NeRF}  & 20.67 & .8561 & .2306 & 24.95 & .8323 & .1962 & 31.58 & .9534 & .1634\\
        See-NeRF\textsuperscript{EvD}        & \Checkmark      & eCRF~\cite{evdeblurnerf} & ours    & 22.02 & .8669 & .2474 & 16.43 & .6932 & .3210 & 29.52 & .9267 & .2776 \\
        See-NeRF        & \Checkmark    & ours     & ours   & \textbf{24.13} & \textbf{.8927} & \textbf{.1916} & \textbf{26.49} & \textbf{.8953} & \textbf{.1638} & \textbf{32.70} & \textbf{.9564} & \textbf{.1574} \\
        \bottomrule
        \end{tabular}
    \label{tab:3}
    \vspace{-1em}
\end{table*}

\subsection{Ablation study}
\label{sec:5.4}

\myPara{Effect of RGB CRF model:}
Comparing See-NeRF\textsuperscript{Base} (training the base model without events) and See-NeRF\textsuperscript{NoEv} (training See-NeRF without events) in \cref{tab:3} shows that our RGB CRF model enables the network to perform HDR NVS while improving the deblurring NVS results.

\begin{figure}[t]
    \centering
    \includegraphics[width=0.99\linewidth]{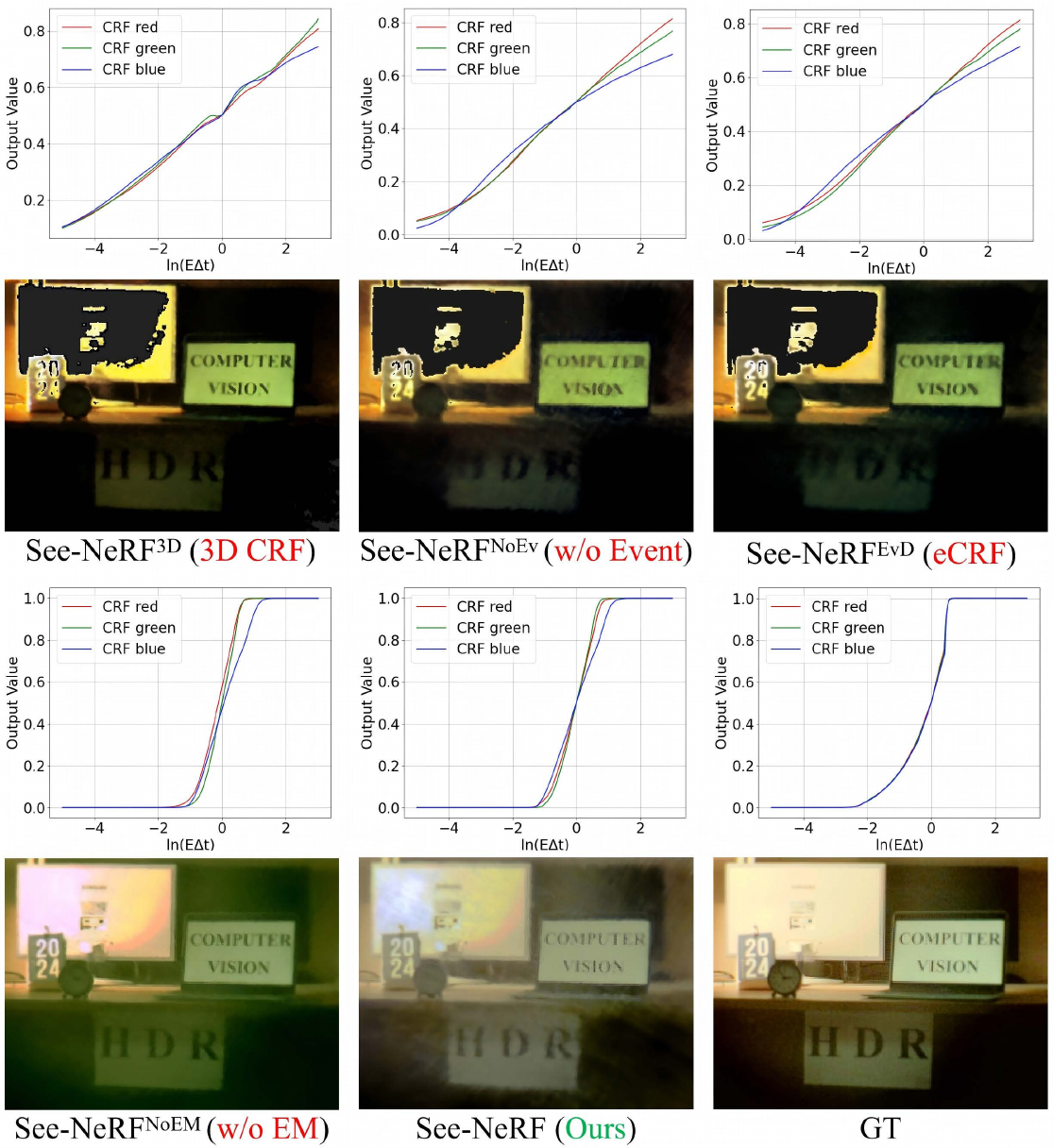}
    \caption{Qualitative results (estimated CRF curve and HDR NVS results) on our real ``computerhard'' scene for the ablation study.}
    \label{fig:8}
\end{figure}

\myPara{2D pixel-wise CRF versus 3D points CRF:}
The results of See-NeRF\textsuperscript{3D} of \cref{tab:3} show that substituting our 2D pixel-wise RGB CRF model with the 3D point CRF of~\cite{hdrnerf, hdr-gs, Evhdr-NeRF} significantly reduces performance in HDR and deblurring NVS and leads to incorrect CRF curve estimation in \cref{fig:8}.

\myPara{Effect of event input:}
The second and third lines of \cref{tab:3} prove that event supervision plays a key role in both HDR and deblurring NVS, as theoretically deduced in \cref{sec:3}.
\cref{fig:8} further indicates that events can help our model to learn the correct CRF curve of the RGB sensor.

\myPara{Effect of event CRF model:}
Comparing See-NeRF\textsuperscript{NoEM} and See-NeRF in \cref{tab:3}, our event CRF model marginally helps align the CRF curve (\cref{fig:8}), achieving further improvement on HDR and Deblurring NVS.
However, replacing our event CRF model with the eCRF in EvDeblur-NeRF~\cite{evdeblurnerf} leads to barely satisfactory reconstruction results and CRF estimation (See-NeRF\textsuperscript{EvD} in \cref{tab:3} and \cref{fig:8}).

\begin{table}[t]
  \centering
  \caption{Discussion on Robustness of See-NeRF.}
  \footnotesize
  \scriptsize
  \setlength{\tabcolsep}{0.5mm}
  \begin{tabular}{c|ccc||c|ccc}
        \toprule
       \multicolumn{4}{c||}{Robustness to Different $\Delta t$ Input}    & \multicolumn{4}{c}{Robustness to Event Noise}  \\
        \midrule
        Single-Exposure & PSNR\textuparrow & SSIM\textuparrow & LPIPS\textdownarrow & Noise Level & PSNR\textuparrow & SSIM\textuparrow & LPIPS\textdownarrow \\
        \midrule
        $\Delta t_0=0.001s$ & 24.66 & .8886 & .2547 & $\times 1$ & \textbf{25.08} & \textbf{.8956} & \textbf{.2324} \\ 
        $\Delta t_2=0.016s$ & 25.08 & .8956 & .2324 & $\times 2$ & 24.99 & \textbf{.8956} & .2342 \\
        $\Delta t_4=0.256s$  & \textbf{25.68} & \textbf{.9062} & \textbf{.2193} & $\times 4$ & 24.86 & .8948 & .2343 \\
        \bottomrule
  \end{tabular}
  \label{tab:4}
  \vspace{-1em}
\end{table}

\subsection{Discussions}
\label{sec:5.5}
\myPara{Robustness to LDR images with different exposures and event noise:}
As shown in \cref{tab:4}, we conduct a robustness experiment on the synthetic ``Catroom'' scene for the HDR NVS task.
Firstly, we use the single-exposure images with exposure times $\Delta t_0, \Delta t_2, \Delta t_4$ as inputs, respectively.
The overall results do not fluctuate significantly, while longer exposure input leads to better results, indicating that the low-light regions of the scene are more difficult to learn than the highlight regions.
Secondly, we use events with different noise levels as input.
\cref{tab:4} and the supplement video results on real data show that event noise does not lead to a significant performance degradation, because NeRF training refers to multi-view input, which can inherently weaken the impact of sparse random noise in each individual view, as proved in RawNeRF~\cite{rawnerf}.
Therefore, See-NeRF does not target the processing of event noise, and it could be a future optimization direction of ERGB-based NVS methods.

\begin{table}[ht]
  \centering
  \caption{Discussion on training and rendering time and 3DGS.}
  \footnotesize
  \scriptsize
  \setlength{\tabcolsep}{0.2mm}
  \begin{tabular}{c||ccc|ccc|c|c}
        \toprule
       & \multicolumn{3}{c|}{Novel Exposure}    & \multicolumn{3}{c|}{HDR}  & Training & Rendering\\
        \midrule
        & PSNR\textuparrow & SSIM\textuparrow & LPIPS\textdownarrow & PSNR\textuparrow & SSIM\textuparrow & LPIPS\textdownarrow & hours & fps \\
        \midrule
        HDR-GS~\cite{hdr-gs} & 19.19 & .7745 & .2537 & 9.25 & .4036 & .6376 & 0.03 & 4.29\\ 
        Ours + 3DGS & 20.39 & .8204 & .2166 & 18.27 & .7244 & .3481 & 0.05 & 5.11\\
        Ours & \textbf{24.15} & \textbf{.8576} & \textbf{.1154} & \textbf{19.43} & \textbf{.7979} & \textbf{.2469} & 2.5 & 0.37 \\
        \bottomrule
  \end{tabular}
  \label{tab:5}
  \vspace{-1em}
\end{table}

\myPara{Training and rendering time and availability on 3DGS:}
The training time of our See-NeRF (2.5 h) is slightly longer than that of HDR-NeRF (2.1 h) due to the introduction of event-related computation. 
We also replace the NeRF network of our method with 3DGS and conduct an experiment on the ``computerhard'' scene of our real data.
As shown in \cref{tab:5}, ``Ours + 3DGS'' is better than HDR-GS, especially on the HDR NVS task, and the training speed is also increased by about 50 times with the high-speed rendering of 3DGS.
Note that the rendering speed of ``Ours + 3DGS'' is even faster than HDR-GS, demonstrating our 2D pixel-wise RGB CRF model is more efficient than the 3D point CRF of HDR-GS.
However, the performance of ``Ours + 3DGS'' is significantly lower than See-NeRF on the novel exposure task.
We think this is because the MLP-based CRF model does not perfectly match the 3DGS during optimization, and ERGB-based HDR 3DGS may be a promising future work.


\section{Conclusion}
\label{sec:6}
In this paper, we propose a unified sensor-physics grounded framework, See-NeRF, for deblurring HDR NVS from single-exposure LDR images and corresponding events, which bridges the discrepancy between the scene radiance and the sensor-recorded ERGB data based on the physical imaging laws and the sensor processing characteristics, achieving the state-of-the-art deblurring HDR NVS results.


\section*{Acknowledgements}
This work is partially supported by grants from the National Natural Science Foundation of China (No. 62132002), Guizhou Provincial Major Scientific and Technological Program (Qiankehe Zhongda [2025] No. 032), Beijing Nova Program (No.20250484786), the Fundamental Research Funds for the Central Universities, and the China Scholarship Council (No. 202506020106)

{
    \small
    \bibliographystyle{ieeenat_fullname}
    \bibliography{main}
}

\renewcommand\thesection{\Alph{section}}
\setcounter{section}{0}
\setcounter{equation}{15}
\setcounter{figure}{8}
\setcounter{table}{5}


\section{Details of event calibration and division}
\label{sec:A}

\myPara{Photometric quantity calibration function $h(\cdot)$:}
For the input event bin $B(t_{i},t_{i+1})$ between $t_{i}$ and $t_{i+1}$, we use $(x,y,t_{\text{first}}^{i},p_{\text{first}}^{i})$ and $(x,y,t_{\text{last}}^{i},p_{\text{last}}^{i})$ to represent the first and last triggered event at pixel $(x,y)$, respectively.
Then we can calculate the offset of each pixel for the input event bin $B(t_{i},t_{i+1})$ with:
\begin{equation}
\begin{split}
h(B(t_{i}, t_{i+1})) =
& \phi(\frac{(t_{i+1}-t_{\text{last}}^{i})}{t_{\text{first}}^{i+1}-t_{\text{last}}^{i}},{p_{\text{first}}^{i+1}}) \\
& -\phi(\frac{(t_i-t_{\text{last}}^{i-1})}{t_{\text{first}}^{i}-t_{\text{last}}^{i-1}},{p_{\text{first}}^{i}}),
\end{split}
\label{eq:16}
\end{equation}
where $\phi(\cdot)$ is a conditional function to determine whether there are adjacent events triggered before and after the current event bin, and whether the polarities are consistent:
\begin{equation}
    \phi(z, p_{\text{first}}^{i})=
    \begin{cases}
    zp_{\text{first}}^{i} & \text{if } p_{\text{first}}^{i} = p_{\text{last}}^{i-1} \\
    0 & \text{if } p_{\text{first}}^{i} \neq p_{\text{last}}^{i-1} \\
    0.5 & \text{if } \nexists p_{\text{first}}^{i} \lor \nexists p_{\text{last}}^{i-1}
\end{cases}
\label{eq:17}
\end{equation}
Since the above operations are the same for all pixels, we omit the $(x,y)$ in \cref{eq:16} and \cref{eq:17}.

\myPara{Events division:}
For each training view, we split the input event stream $B(t_{\text{start}},t_{\text{end}})$ that corresponding to the input LDR blurry image $\mathcal{I}_{\text{LDR}}$ into $b$ bins $B(t_{i},t_{i+1})\}_{i=0}^{b-1}$, where $t_0=t_{\text{start}}$ and $t_{b}=t_{\text{end}}$ are the start and end exposure time of the image. $t_1,...,t_{b-1}$ are the time points that divide the event stream into $b$ event bins with an equal number of events as in the E\textsuperscript{3}NeRF~\cite{e3nerf}, which leverages the temporal blur prior in event distribution for event loss optimization, achieving better performance.


\section{Details of our dataset generation}
\label{sec:B}

\subsection{Synthetic dataset:}
\label{sec:B.1}

Our synthetic dataset consists of 8 HDR Blender~\cite{Blender} scenes (``bathroom'', ``catroom'', ``dogroom'', ``diningroom'', ``sofa'', ``sponza'', ``toyroom'', ``warmroom'') in the synthetic data of HDR-NeRF~\cite{hdrnerf}, which contain both highlight and darkness areas in each scene, resulting in the rendered single-exposure images containing both overexposure and underexposure areas as in the first column of \cref{fig:sup4}.
All synthetic data are at a resolution of $400 \times 400$. 
Below, we introduce the training and test data generation process, respectively.

\myPara{Training images and events generation:}
For each scene, there are 18 views of LDR blurry images with a single-exposure time $\Delta t_2$ and corresponding events within the exposure time for training.
For each view, we add camera shaking and generate 17 sharp HDR raw images $\{\mathcal{I}_{\text{HDR}}^{i}\}_{i=0}^{16}$ at time $t_i$ with camera poses $p(t_i)$.
The camera pose $p(t_i)$ changes over time, indicating the camera shake process.
Note that $t_{\text{start}}\leq t_i\leq t_{\text{end}}$, where $t_{\text{start}}$ and $t_{\text{end}}$ are the start time and end time of the exposure as defined in the main paper.
Then the 17 sharp HDR raw images are averaged to obtain the blurred HDR raw image $\mathcal{I}_{\text{HDR}}^{\text{blur}}$.
We used the classic Reinhard tone mapping~\cite{ReinhardTonemapping} algorithm in \cref{eq:tonemapping} to convert the blurred HDR raw image into an LDR blurred image with exposure time $\Delta t_{2}$ for training as in HDR-NeRF~\cite{hdrnerf}:
\begin{equation}
    \mathcal{I}_{\text{LDR}}^{\Delta t_{2}} = (\frac{\Phi\Delta{t_{2}}{\mathcal{I}}_{\text{HDR}}^{\text{blur}}}{\Phi\Delta{t_{2}}{\mathcal{I}}_{\text{HDR}}^{\text{blur}}+1})^{\frac{1}{2.2}}.
    \label{eq:tonemapping}
\end{equation}
$\Phi$ is a scale factor, and we take it as 62.5.
The 17 sharp HDR raw images $\{\mathcal{I}_{\text{HDR}}^{i}\}_{i=0}^{16}$ are also input into the event simulator v2e~\cite{v2e} with ``noisy'' option, which involves the event latency and noise simulation into the data generation.
Then the events corresponding to the synthetic LDR blurry image $\mathcal{I}_{\text{LDR}}^{\Delta t_{2}}$ is obtained:
\begin{equation}
    B(t_{\text{start}}, t_{\text{end}}) = \mathbf{v2e}(\{\mathcal{I}_{\text{HDR}}^{i}\}_{i=0}^{16}).
    \label{eq:18}
\end{equation}
Note that we add a Bayer filter in the v2e to generate the color events as in the DAVIS 346 Color~\cite{davis346} event camera.

\myPara{Additional training images:}
For each training view, we also use \cref{eq:tonemapping} to tone map the 8-th raw sharp image $\mathcal{I}_{\text{HDR}}^{7}$ in $\{\mathcal{I}_{\text{HDR}}^{i}\}_{i=0}^{16}$ into LDR domain with exposure times $\Delta t_{0}$, $\Delta t_{2}$ and $\Delta t_{4}$, which are used to train the RGB-based HDR NVS works~\cite{hdrnerf,hdr-gs,gausshdr,gaussian-in-the-dark} without deblurring effects and to train HDR-NeRF~\cite{hdrnerf} with multi-exposure sharp LDR images as input for reference.

\myPara{Test images generation:}
For each scene, there are 17 novel views for testing.
For each test view, we render 1 ground-truth (GT) sharp HDR raw image $\mathcal{I}_{\text{HDR}}^{\text{sharp}}$ with Blender for the HDR novel view synthesis (NVS) test.
We also use \cref{eq:tonemapping} to transform the sharp HDR image $\mathcal{I}_{\text{HDR}}^{\text{sharp}}$ into two GT sharp LDR images with novel exposure times $\Delta t_{1}$ and $\Delta t_{3}$ for the novel exposure NVS test.

\subsection{Real dataset:}
\label{sec:B.2}

Our real dataset consists of 5 real-world extreme lighting scenes (``bear'', ``computer'', ``computerhard'', ``desktop'', ``table'') with both highlight and darkness areas, resulting in the captured single-exposure LDR images containing both overexposure and underexposure areas.
The upper part of \cref{tab:sup1} shows the minimum and maximum luminance of the scenes in our real data, which contains a very wide dynamic range.
All data is obtained by a DAVIS 346 Color camera~\cite{davis346}, which can capture spatial-temporal aligned images and corresponding events at resolution $346 \times 260$.

\myPara{Training images and events generation:}
For each scene, there are 16 training views.
For each view, we capture 1 LDR blurry image at a single-exposure time $\Delta t_2$ and corresponding events within the exposure time with a handheld  DAVIS 346 Color camera.

\myPara{Additional training images:}
For each training view, we also capture three sharp LDR images with exposure times $\Delta t_{0}$, $\Delta t_{2}$, and $\Delta t_{4}$ with a tripod-fixed DAIVS 346 Color camera to train the RGB-based HDR NVS works~\cite{hdrnerf,hdr-gs,gausshdr,gaussian-in-the-dark} without deblurring effects and to train HDR-NeRF~\cite{hdrnerf} with multi-exposure sharp LDR images as input for reference.

\myPara{Test images generation:}
For each scene, there are 12 novel views for testing.
For each test view, we capture 5 LDR sharp images with exposure times $\{\Delta t_i\}_{i=0}^4$ with a tripod-fixed DAIVS 346 Color camera.
The LDR sharp images with novel exposure times $\Delta t_1$ and $\Delta t_3$ are used for the novel exposure NVS test, which is the same as the synthetic dataset.
The GT HDR sharp image for the HDR NVS test is obtained by merging the five LDR sharp images with exposure times $\{\Delta t_i\}_{i=0}^4$ with the classic multi-exposure HDR reconstruction algorithm Debevec~\cite{hdr01} as in RawHDR~\cite{single-raw-hdr}.


\begin{table}[t]
    \caption{The illumination conditions of our real-world scenes and the exposure time settings of our synthetic and real-world datasets.}
    \centering
    \scriptsize
    \setlength{\tabcolsep}{0.1mm}
    \begin{tabular}{c|c|c|c|c|c|c}
    \toprule
    & Synthetic & \multicolumn{5}{c}{Real-World Scenes} \\
    & Scenes & {Bear}  & {Computer} & {Computerhard} & {Desktop} &{Table} \\
    
    \midrule
    Min Scene luminance & - & 70 lux   & 50 lux    & 30 lux    & 20 lux   & 10 lux\\
    Max Scene luminance & - & 2000 lux & 4000 lux  & 5000 lux  & 1000 lux & 1500 lux\\
    
    \midrule
    Exposure Time $\Delta{t}_0$ & 0.001 s & 0.05 s & 0.02 s & 0.002 s & 0.05 s & 0.05 s \\
    Exposure Time $\Delta{t}_1$ & 0.004 s & 0.10 s & 0.04 s & 0.004 s & 0.10 s & 0.10 s \\
    Exposure Time $\Delta{t}_2$ & 0.016 s & 0.15 s & 0.06 s & 0.006 s & 0.15 s & 0.15 s \\
    Exposure Time $\Delta{t}_3$ & 0.064 s & 0.20 s & 0.08 s & 0.008 s & 0.20 s & 0.20 s \\
    Exposure Time $\Delta{t}_4$ & 0.256 s & 0.25 s & 0.10 s & 0.010 s & 0.25 s & 0.25 s \\

    \bottomrule
    \end{tabular}
    \label{tab:sup1}
    \vspace{-1em}
\end{table}

\section{Implementation details}
\label{sec:C}

Our framework is built upon E\textsuperscript{2}NeRF~\cite{e2nerf}, and all experiments are conducted on a single RTX 3090 GPU.
We set the number of sampled points on each ray to 64 and 128 for the coarse and fine networks, respectively.
The batch size of sampled rays is set to 512 and 1024 for the real data and synthetic data, respectively, and we take 50k iterations for each scene.
The above NeRF network settings are the same for all the compared NeRF-based methods and our See-NeRF in the experiments.
For the 3DGS-based methods, the number of iterations is set to 30000, and other parameters are set to their default parameter.
For the detailed evaluation of hyperparameters $b$, $\lambda$, and training time of our See-NeRF, please refer to \cref{sec:D.2}.

\subsection{Exposure time settings}
\label{sec:C.1}
The lower part of \cref{tab:sup1} shows the value of exposure times $\{\Delta t_i\}_{i=0}^4$ of our synthetic and real datasets.
For the Real-World-Challenge dataset~\cite{e3nerf} of the deblurring experiment, the exposure times of the input blurry images are 0.1 s, 0.12 s, 0.1 s, 0.1 s, 0.15 s, for the ``corridor'', ``lab'', ``lobby'', ``shelf'', ``table'' scenes, respectively.
We use these exposure times to train and test our See-NeRF to obtain the corresponding deblurred novel view LDR images.

\subsection{Pose estimation}
\label{sec:C.2}
We use the pose generation framework of E\textsuperscript{2}NeRF~\cite{e2nerf} to obtain the poses $\{p(t_i)\}_{i=0}^{b}$ corresponding to the blurry images for all the compared methods and our See-NeRF, which ensures the input poses are the same, so the reconstruction result is only affected by the network architecture.
Recent RGB-based deblurring NVS works provide different camera pose estimation strategies~\cite{bad-nerf, bad-gs, dpnerf} and dynamic scene pose estimation strategies~\cite{moblurf, crim-gs}.
Integrating the event supervision into these methods would be a promising future research direction for the ERGB-based HDR deblurring NVS task, as demonstrated in EBAD-NeRF~\cite{ebad-nerf}.

\subsection{Evaluation of HDR results}
\label{sec:C.3}
For the HDR NVS task, we first use the ``Enhancer'' and ``Compressor'' options of Photomatrix Pro~\cite{Photomatix} to tone map the reconstructed and ground truth HDR images into the LDR domain for the synthetic and real data, respectively.
Then, we use the tone-mapped images to quantitatively and qualitatively evaluate our results in our main paper and supplementary materials, as in~\cite{hdr-gs,hdrnerf}.


\begin{figure*}[t]
  \centering
  \includegraphics[width=1\linewidth]{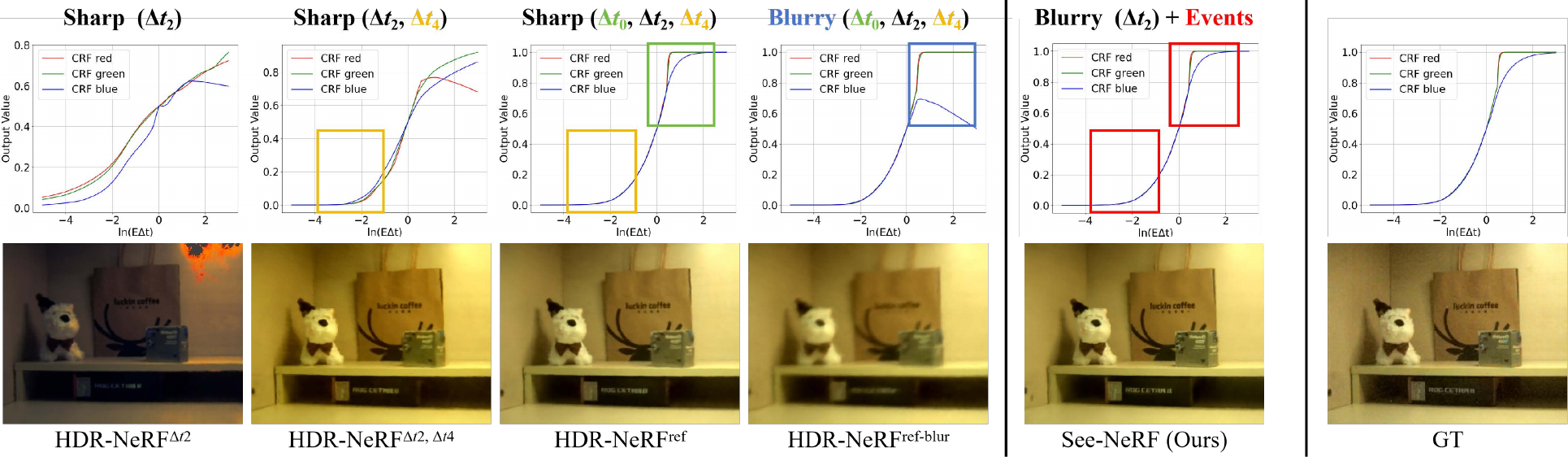}
   \caption{Evaluation of the HDR NVS performance boundaries between multi-exposure RGB-based HDR-NeRF and our single-exposure ERGB-based See-NeRF on the real ``Table'' scene. The upper and lower parts of the figure are the estimated CRF curves and HDR NVS results, respectively.
   See-NeRF can effectively leverage events to replace the CRF curve correction effect of the LDR images with exposure times $\Delta t_0$ and $\Delta t_4$ in HDR-NeRF\textsuperscript{ref}. Besides, See-NeRF is also robust to the image motion blur compared to HDR-NeRF\textsuperscript{ref-blur}.
   }
   \label{fig:sup1}
\end{figure*}

\section{Additional evaluations}

\label{sec:D}

\subsection{Single-exposure vs. multi-exposure}
\label{sec:D.1}
In this experiment, we train HDR-NeRF~\cite{hdrnerf} with different input exposure ranges to evaluate the HDR NVS performance boundaries of multi-exposure RGB-based HDR-NeRF and our single-exposure ERGB-based See-NeRF.

\begin{table}[t]
    \caption{Evaluation of the HDR NVS performance boundaries between multi-exposure RGB-based HDR-NeRF and our single-exposure ERGB-based See-NeRF on the real ``Table'' scene.}
    \centering
    \scriptsize
    \setlength{\tabcolsep}{0.6mm}

    \begin{tabular}{l|ccc|ccc}
    \toprule
    & \multicolumn{6}{c}{Real ``Table'' Scene of Our Dataset} \\
    
    & \multicolumn{3}{c|}{Input Data} & \multicolumn{3}{c}{HDR Results}\\
    & Exposure Time & Image & Event & PSNR\textuparrow & SSIM\textuparrow & LPIPS\textdownarrow \\
    
    \midrule
    HDR-NeRF\textsuperscript{ref}      & $\Delta t_0, \Delta t_2, \Delta t_4$ & Sharp & - & 28.23 & .9207 & .1550 \\
    HDR-NeRF\textsuperscript{$\Delta t_2, \Delta t_4$}    & $\Delta t_2, \Delta t_4$ & Sharp & - & 21.78 & .8404 & .1867 \\
    HDR-NeRF\textsuperscript{$\Delta t_2$}                           & $\Delta t_2$  & Sharp & - & 12.47 & .6804 & .3551 \\
    
    \midrule
    HDR-NeRF\textsuperscript{ref-blur} & $\Delta t_0, \Delta t_2, \Delta t_4$ & Blurry & - & 25.54 & .8640 & .3017 \\
    See-NeRF                           & $\Delta t_2$  & Blurry & \Checkmark  & \textbf{27.12} & \textbf{.9093} & \textbf{.1548} \\

    \bottomrule
    \end{tabular}
    \label{tab:sup2}
\end{table}

\myPara{Quantitative results:}
HDR-NeRF\textsuperscript{ref} in \cref{tab:sup2} demonstrates that multi-exposure RGB-based HDR-NeRF only surpasses our single-exposure ERGB-Based See-NeRF in overall performance when the input LDR images are sharp and cover exposure times at all three scales: short ($\Delta t_0$), medium ($\Delta t_2$), and long ($\Delta t_4$).
When the input images are motion-blurred, the results of HDR-NeRF\textsuperscript{ref-blur} will be significantly affected by the motion blur and degraded.
In contrast, our See-NeRF achieves the best HDR NVS results with blurry single-exposure LDR images and corresponding events.

\myPara{Qualitative results:}
HDR-NeRF\textsuperscript{$\Delta t_2$} in \cref {fig:sup1} illustrates that with images with single-exposure time $\Delta t_2$ as input, the estimated CRF curve deviates completely from GT, and the HDR NVS results are also completely distorted.
When adding the images with longer exposure time $\Delta t_4$ for training, HDR-NeRF\textsuperscript{$\Delta t_2, \Delta t_4$} can accurately recover the CRF curve of the low-light area.
At the same time, there is still a deviation in the CRF curve of the highlight area, and the HDR NVS results will show a color cast in the highlight area.
When further adding the images with shorter exposure time $\Delta t_0$, the CRF curve of the highlight area is further corrected and closer to the GT CRF curve, as shown in the results of HDR-NeRF\textsuperscript{ref}.
However, in the case of blurry input, the estimated CRF curves of HDR-NeRF\textsuperscript{ref-blur} are distorted, and the reconstruction results are blurred.
In comparison, See-NeRF achieves accurate CRF curve estimation and produces sharp, high-quality HDR NVS results.
This demonstrates that See-NeRF effectively leverages spatial and temporal differential information of events in low-light, highlight, and blurry regions to infer the actual scene brightness to enhance CRF estimation and eliminate the effects of blurred input, which is consistent with the theoretical proof in Sec. 3.

\begin{table}[h]
  \centering
  \caption{Evaluation of parameter $b$ and training time.}
  \footnotesize
  \scriptsize
  \setlength{\tabcolsep}{1.7mm}
  \begin{tabular}{c||ccc|ccc|c}
        \toprule
       & \multicolumn{3}{c|}{HDR (``Desktop'' scene)}    & \multicolumn{3}{c|}{Deblurring (``Lab'' scene)}  & Time\\
        \midrule
        $b$ & PSNR\textuparrow & SSIM\textuparrow & LPIPS\textdownarrow & PSNR\textuparrow & SSIM\textuparrow & LPIPS\textdownarrow & hours\\
        \midrule
        2 & 27.10 & .9127 & .1930 & 33.98 & .9657 & .1458 & 1.75 \\ 
        4 & 28.13 & .9152 & .1796 & 34.56 & .9681 & .1374 & 2.50 \\
        6 & 26.89 & .8856 & .1842 & 35.21 & .9712 & .1277 & 3.23 \\
        8 & 26.49 & .8753 & .1834 & 35.57 & .9742 & .1193 & 3.95 \\
        \bottomrule
  \end{tabular}
  \label{tab:sup3}
\end{table}

\begin{table*}[t]
    \centering
    \caption{Results of the ablation study on previous related CRF models. The results are the averages of all scenes. LPF: Low-Pass Filter}
    \scriptsize
    \setlength{\tabcolsep}{1.2mm}
        \begin{tabular}{l|cc|ccc|ccc}
        \toprule
        & 
        & 
        & \multicolumn{3}{c|}{Our Real Data (HDR)}
        & \multicolumn{3}{c}{Real-World-Challenge Data} \\

        & Event CRF Model & RGB CRF Model & PSNR\textuparrow & SSIM\textuparrow & LPIPS\textdownarrow & PSNR\textuparrow & SSIM\textuparrow & LPIPS\textdownarrow  \\
   
        \midrule
        See-NeRF\textsuperscript{Pow}    & ours    & Power Function $\mathcal{F}(x)=x^a$ CRF~\cite{evhdr-gs} & 19.75 & .7239 & .2696 & 28.35 & .9242 & .2657 \\
        
        See-NeRF                      & ours & MLP-Based CRF (ours) & \textbf{26.49} & \textbf{.8953} & \textbf{.1638} & \textbf{32.70} & \textbf{.9564} & \textbf{.1574} \\
        
        \midrule
        See-NeRF\textsuperscript{v2e}   & Explicitly Parameterized 2nd-Order LPF~\cite{v2e} & ours & 22.10 & .8644 & .1938 & 30.51 & .9441 & .1748 \\
        
        See-NeRF\textsuperscript{Deblur \emph{e}} & Explicitly Parameterized 4th-Order LPF~\cite{deblurenerf} & ours & 24.28 & 8859 & .1847 & 31.84 & .9521 & .1664 \\
        
        See-NeRF                      & Implicitly MLP-Optimized 2nd-Order LPF (ours) & ours & \textbf{26.49} & \textbf{.8953} & \textbf{.1638} & \textbf{32.70} & \textbf{.9564} & \textbf{.1574} \\
        \bottomrule
        \end{tabular}
    \label{tab:9}
\end{table*}

\subsection{Parameters and efficiency}
\label{sec:D.2}
\myPara{Parameter $b$:}
$b$ in Eq.~(6) represents the discretization of Eq.~(1) with $b$ discrete time points.
Therefore, a larger $b$ leads to better simulation of the imaging process and generates better results, but at the same time requires more iterations and increases the training time as shown in \cref{tab:sup3}.
For the HDR NVS task, when $b=4$, See-NeRF gets the best results.
For the deblurring NVS task, when $b>6$, the improvement is marginal.
So we set $b=4$ for the HDR and novel exposure NVS experiments on our proposed real-world datasets and $b=6$ for the deblurring NVS experiments for the Real-World-Challenge dataset.

\myPara{Training and rendering time:}
Compared to HDR-NeRF (2.1h), See-NeRF (2.5h) increases computation and training time by 19.05\% under the same conditions (same NeRF-related parameters, number of iterations, and batch size; on a single RTX 3090 GPU; on our real dataset).
The increased computation is mainly due to the event-related calculations.
However, the rendering speed of See-NeRF (0.37 fps) is slightly faster than HDR-NeRF (0.35 fps) under a resolution of 346*260.
This is mainly because no events are involved during test time, and our 2D pixel-wise tone mapping is more efficient than the 3D points tone mapping of HDR-NeRF.

\begin{figure}[h]
  \centering
  \includegraphics[width=1\linewidth]{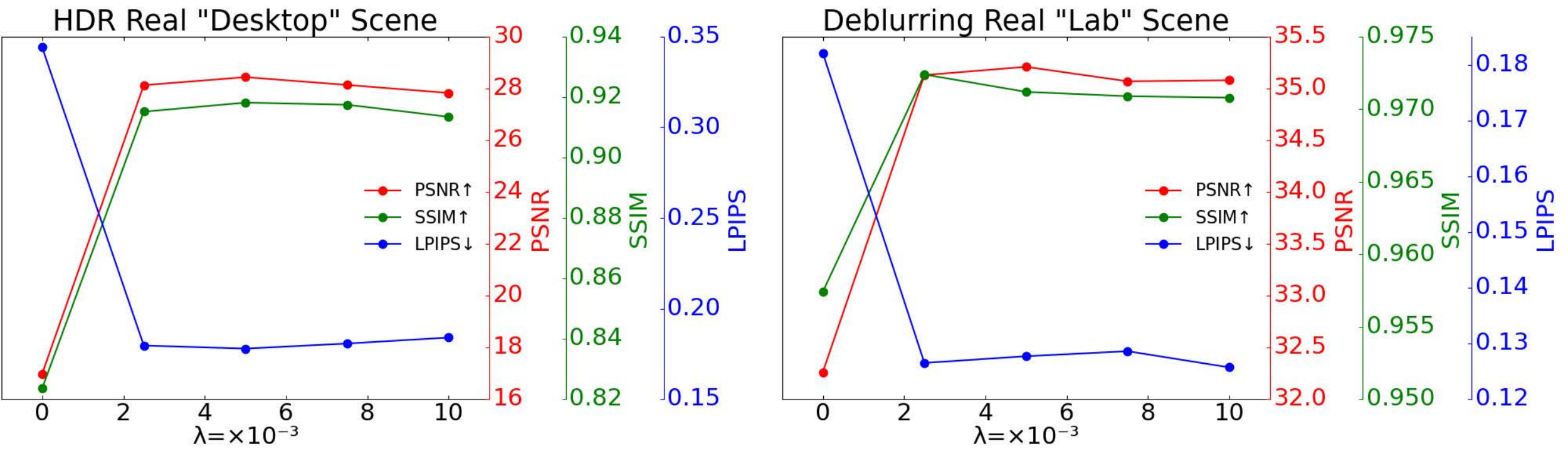}
   \caption{Evaluation of parameter $\lambda$.}
   \label{fig:sup2}
\end{figure}

\myPara{Parameter $\lambda$:}
As in \cref{fig:sup2}, we evaluate the impact of $\lambda$ on the real-world ``desktop'' and ``lab'' scenes for the HDR and deblurring NVS tasks, respectively.
When $\lambda=0.005$, See-NeRF can achieve the best overall results on both tasks.

\begin{figure}[t]
  \centering
  \includegraphics[width=1\linewidth]{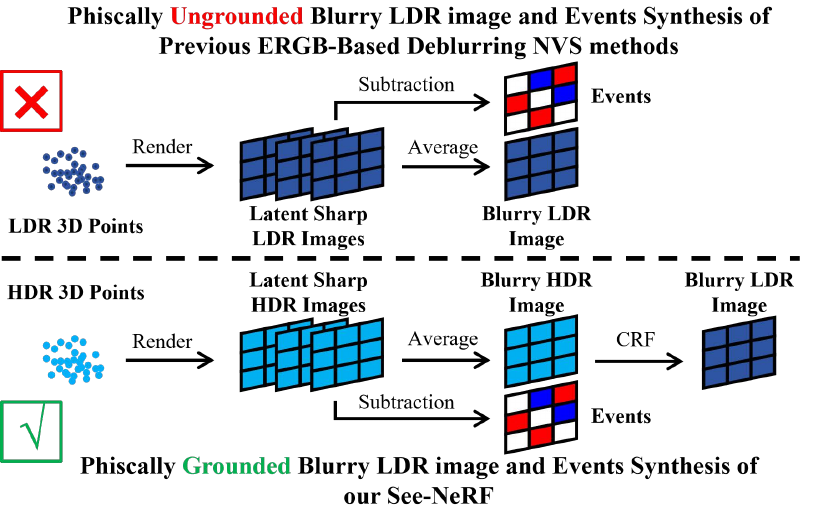}
   \caption{Analysis of sensor-physics grounded deblurring effect.}
   \label{fig:sup3}
\end{figure}

\subsection{Sensor-Physics grounded deblurring effect}
\label{sec:D.3}
Our See-NeRF significantly surpasses previous ERGB-based deblurring NVS methods E\textsuperscript{2}NeRF~\cite{e2nerf}, E\textsuperscript{3}NeRF~\cite{e3nerf}, E2GS~\cite{e2gs}, and Ev-DeblurNeRF~\cite{evdeblurnerf} as shown in \cref{tab:sup10}.
This is because these methods all perform the motion blur and events synthesis directly on the nonlinear LDR images rendered by their 3D representation network, as shown in \cref{fig:sup3}.
In contrast, our See-NeRF performs the motion blur and events synthesis on the linear raw HDR values of the scene radiance rendered by the NeRF network, and then tone maps the blurry raw HDR image to the LDR domain, aligning with the physical imaging process of sensors in the real world.
On the other hand, our event CRF model eliminates the effects of event delay and further improves the performance.
Therefore, See-NeRF achieves SOTA results in the ERGB-based deblurring NeRF task.

\subsection{Ablation on previous related CRF models}
\label{sec:D.4}
We conduct an ablation study on previous related CRF models on our real-world dataset and the Real-World-Challenge dataset, respectively, as shown in \cref{tab:9}

\myPara{RGB CRF model of EvHDR-GS~\cite{evhdr-gs}:}
Recently, EvHDR-GS uses 3DGS~\cite{3DGS} as a backbone for the ERGB-based HDR video reconstruction task.
Though it uses a similar design that RGB CRF is applied after volumetric rendering, it does not consider the image integration process and simply models the RGB CRF with one power function $\mathcal{F}(x)=x^a$ for the three color channels during the 2D tone mapping, while See-NeRF address these two aspects with Eq.~(6) in the main paper and a three-independent MLP-Based RGB CRF model, as shown in the Fig.~3 of the main paper.
Therefore, applying its imaging model to See-NeRF degrades the performance significantly on both HDR and deblurring NVS tasks, as shown in the results of See-NeRF\textsuperscript{Pow} in \cref{tab:9}.

\myPara{Event CRF model of v2e~\cite{v2e} and Deblur \emph{e}-NeRF~\cite{deblurenerf}:}
v2e and Deblur \emph{e}-NeRF models intensity-dependent pixel bandwidth and latency with the explicitly parameterized 2nd-order and 4th-order low-pass filter, respectively.
However, our See-NeRF incorporates scene/camera-related event delay factors into an MLP to calculate the delay coefficient for the 2nd-order low-pass filter.
Replacing our event CRF model with the models of v2e and Deblur \emph{e}-NeRF weakens the performance of See-NeRF on both deblurring and HDR tasks, as in the results of See-NeRF\textsuperscript{v2e} and See-NeRF\textsuperscript{Deblur \emph{e}} in \cref{tab:9}, proving the effectiveness of our event CRF model.

\begin{table*}[t]
    \centering
    \caption{Detailed quantitative novel exposure NVS results on each scene of our synthetic dataset.}
    \label{tab:sup4}
    \scriptsize
    \setlength{\tabcolsep}{0.35mm}
        \begin{tabular}{c|l|cc|ccc|ccc|ccc|ccc}
        
        \toprule
        & & \multicolumn{2}{c|}{Input} 
        & \multicolumn{3}{c}{Bathroom} 
        & \multicolumn{3}{|c}{Catroom} 
        & \multicolumn{3}{|c}{Diningroom} 
        & \multicolumn{3}{|c}{Dogroom}  \\
        
        \midrule
        & Methods & Image & Event & PSNR\textuparrow & SSIM\textuparrow & LPIPS\textdownarrow & PSNR\textuparrow & SSIM\textuparrow & LPIPS\textdownarrow & PSNR\textuparrow & SSIM\textuparrow & LPIPS\textdownarrow & PSNR\textuparrow & SSIM\textuparrow & LPIPS\textdownarrow \\
        
        \midrule
        Reference & HDR-NeRF\textsuperscript{ref} & Sharp ($\Delta t_0$, $\Delta t_2$, $\Delta t_4$) & - & 29.69 & .9148 & .1798 & 26.56 & .9089 & .1589 & 29.02 & .9493 & .1053 & 25.20 & .8537 & .1028 \\

        \midrule
        \multirow{4}{*}{\makecell[c]{RGB-Based \\ HDR NVS}}
        & HDR-NeRF~\cite{hdrnerf}        & Sharp ($\Delta t_2$) & - & 27.13 & .9161 & .1920 & 24.56 & .8954 & .1769 & 23.68 & .9454 & .1002 & 21.11 & .8382 & .1088 \\
        & HDR-GS~\cite{hdr-gs}          & Sharp ($\Delta t_2$) & - & 17.17 & .7536 & .3440 & 15.66 & .7041 & .3088 & 12.73 & .6970 & .4790 & 13.24 & .7189 & .3377 \\
        & Gaussian-DK~\cite{gaussian-in-the-dark} & Sharp ($\Delta t_2$) & - & 29.99 & .9599 & .1580 & 28.46 & .9478 & .1636 & 21.21 & .9142 & .1409 & 24.24 & .9714 & .1593  \\
        & GaussHDR~\cite{gausshdr}          & Sharp ($\Delta t_2$) & -  & 15.30 & .7288 & .1925 & 14.37 & .7367 & .1858 & 12.73 & .7483 & .2230 & 12.03 & .7115 & .2473 \\
        
        \midrule
        \multirow{3}{*}{\makecell[c]{ERGB-Based \\ Deblurring \\ HDR NVS}}
        & HDR-NeRF+       & Blurry ($\Delta t_2$) & \Checkmark & 19.50 & .7311 & .3197 & 18.58 & .7288 & .2970 & 14.59 & .7336 & .2699 & 14.74 & .5893 & .3586 \\
        & EvHDR-NeRF~\cite{Evhdr-NeRF}      & Blurry ($\Delta t_2$) & \Checkmark & 26.43 & .8729 & .3124 & 25.11 & .8855 & .3249 & 20.45 & .8412 & .2847 & 22.81 & .8229 & .2903  \\
        
        & See-NeRF        & Blurry ($\Delta t_2$) & \Checkmark & \textbf{28.13} & \textbf{.9141} & \textbf{.1596} & \textbf{27.45} & \textbf{.9214} & \textbf{.1774} & \textbf{23.28} & \textbf{.9315} & \textbf{.1314} & \textbf{24.81} & \textbf{.8678} & \textbf{.1430}  \\

        \bottomrule
    \end{tabular}
    \vspace{-1em}
\end{table*}

\begin{table*}[t]
    \centering
    \caption{Detailed quantitative novel exposure NVS results on each scene of our synthetic dataset.}
    \label{tab:sup5}
    \scriptsize
    \setlength{\tabcolsep}{0.35mm}
        \begin{tabular}{c|l|cc|ccc|ccc|ccc|ccc}
        
        \toprule
        & & \multicolumn{2}{c|}{Input} 
        & \multicolumn{3}{c}{Sofa} 
        & \multicolumn{3}{|c}{Sponza} 
        & \multicolumn{3}{|c}{Toyroom} 
        & \multicolumn{3}{|c}{Warmroom}  \\
        
        \midrule
        & Methods & Image & Event & PSNR\textuparrow & SSIM\textuparrow & LPIPS\textdownarrow & PSNR\textuparrow & SSIM\textuparrow & LPIPS\textdownarrow & PSNR\textuparrow & SSIM\textuparrow & LPIPS\textdownarrow & PSNR\textuparrow & SSIM\textuparrow & LPIPS\textdownarrow \\
        
        \midrule
        Reference & HDR-NeRF\textsuperscript{ref} & Sharp ($\Delta t_0$, $\Delta t_2$, $\Delta t_4$) & - & 30.66 & .9292 & .1139 & 31.49 & .9555 & .1161 & 31.69 & .9641 & .0694 & 29.47 & .9224 & .1657 \\

        \midrule
        \multirow{4}{*}{\makecell[c]{RGB-Based \\ HDR NVS}}
        & HDR-NeRF~\cite{hdrnerf}        & Sharp ($\Delta t_2$) & - & 20.48 & .9199 & .1248 & 29.48 & .9504 & .1252 & 25.48 & .9569 & .0934 & 24.97 & .9200 & .1583 \\
        & HDR-GS~\cite{hdr-gs}          & Sharp ($\Delta t_2$) & - & 12.97 & .7777 & .2391 & 18.78 & .7352 & .4568 & 15.04 & .7682 & .3198 & 15.42 & .7873 & .3411 \\
        & Gaussian-DK~\cite{gaussian-in-the-dark} & Sharp ($\Delta t_2$) & - & 24.46 & .9787 & .1351 & 20.27 & .6563 & .3864 & 28.63 & .9681 & .1526 & 29.74 & .9609 & .1378 \\
        & GaussHDR~\cite{gausshdr}          & Sharp ($\Delta t_2$) & -  & 13.79 & .7391 & .2252 & 22.34 & .8678 & .2037 & 12.49 & .7607 & .2293 & 14.50 & .7589 & .2164 \\
        
        \midrule
        \multirow{3}{*}{\makecell[c]{ERGB-Based \\ Deblurring \\ HDR NVS}}
        & HDR-NeRF+       & Blurry ($\Delta t_2$) & \Checkmark & 18.22 & .7045 & .2436 & 18.06 & .6032 & .2696 & 17.18 & .8291 & .1948 & 20.15 & .8232 & .2753 \\
        & EvHDR-NeRF~\cite{Evhdr-NeRF}      & Blurry ($\Delta t_2$) & \Checkmark & 24.04 & .9048 & .2283 & 24.40 & .8551 & .2655 & 25.70 & .9142 & .2871 & 26.05 & .8728 & .2850 \\
        
        & See-NeRF        & Blurry ($\Delta t_2$) & \Checkmark & \textbf{28.03} & \textbf{.9399} & \textbf{.1075} & \textbf{33.33} & \textbf{.9693} & \textbf{.0929} & \textbf{28.76} & \textbf{.9533} & \textbf{.0937} & \textbf{26.75} & \textbf{.9185} & \textbf{.1572} \\

        \bottomrule
    \end{tabular}
    \vspace{-1em}
\end{table*}

\begin{table*}[t]
    \centering
    \caption{Detailed quantitative HDR NVS results on each scene of our synthetic dataset.}
    \label{tab:sup6}
    \scriptsize
    \setlength{\tabcolsep}{0.35mm}
        \begin{tabular}{c|l|cc|ccc|ccc|ccc|ccc}
        
        \toprule
        & & \multicolumn{2}{c|}{Input} 
        & \multicolumn{3}{c}{Bathroom} 
        & \multicolumn{3}{|c}{Catroom} 
        & \multicolumn{3}{|c}{Diningroom} 
        & \multicolumn{3}{|c}{Dogroom}  \\
        
        \midrule
        & Methods & Image & Event & PSNR\textuparrow & SSIM\textuparrow & LPIPS\textdownarrow & PSNR\textuparrow & SSIM\textuparrow & LPIPS\textdownarrow & PSNR\textuparrow & SSIM\textuparrow & LPIPS\textdownarrow & PSNR\textuparrow & SSIM\textuparrow & LPIPS\textdownarrow \\
        
        \midrule
        Reference & HDR-NeRF\textsuperscript{ref} & Sharp ($\Delta t_0$, $\Delta t_2$, $\Delta t_4$) & - & 25.96 & .8412 & .2413 & 26.09 & .8914 & .2015 & 25.45 & .9280 & .1432 & 24.65 & .8332 & .1261 \\

        \midrule
        \multirow{4}{*}{\makecell[c]{RGB-Based \\ HDR NVS}}
        & HDR-NeRF~\cite{hdrnerf}        & Sharp ($\Delta t_2$) & - & 19.06 & .7886 & .3384 & 19.84 & .8531 & .2722 & 11.15 & .4079 & .6993 & 14.65 & .6379 & .3291 \\
        & HDR-GS~\cite{hdr-gs}          & Sharp ($\Delta t_2$) & - & 12.16 & .6562 & .5850 & 13.58 & .7421 & .5033 & 12.90 & .6761 & .4857 & 14.56 & .6995 & .4755 \\
        & Gaussian-DK~\cite{gaussian-in-the-dark} & Sharp ($\Delta t_2$) & - & 13.57 & .7088 & .2689 & 12.04 & .7207 & .2380 & 10.71 & .5619 & .4420 & 12.87 & .7302 & .2583 \\
        & GaussHDR~\cite{gausshdr}          & Sharp ($\Delta t_2$) & - & 13.39 & .7976 & .1398 & 10.81 & .7321 & .1784 & 12.27 & .7389 & .1540 & 10.86 & .7789 & .1763 \\
        
        \midrule
        \multirow{3}{*}{\makecell[c]{ERGB-Based \\ Deblurring \\ HDR NVS}}
        & HDR-NeRF+       & Blurry ($\Delta t_2$) & \Checkmark & 16.18 & .6439 & .3847 & 14.29 & .6153 & .3434 & 12.95 & .5215 & .6971 & 16.15 & .5081 & .3807 \\
        & EvHDR-NeRF~\cite{Evhdr-NeRF}      & Blurry ($\Delta t_2$) & \Checkmark & \textbf{22.85} & .7915 & .3624 & 23.56 & .8594 & .3784 & 23.11 & .8841 & .2452 & 17.62 & .7442 & .3509 \\
        
        & See-NeRF        & Blurry ($\Delta t_2$) & \Checkmark & 20.83 & \textbf{.8282} & \textbf{.2801} & \textbf{25.08} & \textbf{.8956} & \textbf{.2324} & \textbf{24.71} & \textbf{.9219} & \textbf{.1414} & \textbf{21.99} & \textbf{.8638} & \textbf{.1757} \\

        \bottomrule
    \end{tabular}
    \vspace{-1em}
\end{table*}

\begin{table*}[t]
    \centering
    \caption{Detailed quantitative HDR NVS results on each scene of our synthetic dataset.}
    \label{tab:sup7}
    \scriptsize
    \setlength{\tabcolsep}{0.35mm}
        \begin{tabular}{c|l|cc|ccc|ccc|ccc|ccc}
        
        \toprule
        & & \multicolumn{2}{c|}{Input} 
        & \multicolumn{3}{c}{Sofa} 
        & \multicolumn{3}{|c}{Sponza} 
        & \multicolumn{3}{|c}{Toyroom} 
        & \multicolumn{3}{|c}{Warmroom}  \\
        
        \midrule
        & Methods & Image & Event & PSNR\textuparrow & SSIM\textuparrow & LPIPS\textdownarrow & PSNR\textuparrow & SSIM\textuparrow & LPIPS\textdownarrow & PSNR\textuparrow & SSIM\textuparrow & LPIPS\textdownarrow & PSNR\textuparrow & SSIM\textuparrow & LPIPS\textdownarrow \\
        
        \midrule
        Reference & HDR-NeRF\textsuperscript{ref} & Sharp ($\Delta t_0$, $\Delta t_2$, $\Delta t_4$) & - & 28.34 & .9000 & .1504 & 26.24 & .8841 & .2189 & 30.66 & .9525 & .0994 & 27.94 & .8948 & .1828 \\

        \midrule
        \multirow{4}{*}{\makecell[c]{RGB-Based \\ HDR NVS}}
        & HDR-NeRF~\cite{hdrnerf}        & Sharp ($\Delta t_2$) & - & 13.45 & .7020 & .3381 & 24.93 & .8736 & .2469 & 14.60 & .8315 & .2605 & 18.50 & .8126 & .2752 \\
        & HDR-GS~\cite{hdr-gs}          & Sharp ($\Delta t_2$) & - & 12.29 & .6713 & .5124 & 17.00 & .7045 & .5966 & 13.30 & .7606 & .4901 & 12.53 & .6837 & .4284 \\
        & Gaussian-DK~\cite{gaussian-in-the-dark} & Sharp ($\Delta t_2$) & - & 15.00 & .7883 & .2829 & 11.20 & .5116 & .5422 & 12.85 & .7376 & .2544 & 13.66 & .7014 & .2782 \\
        & GaussHDR~\cite{gausshdr}          & Sharp ($\Delta t_2$) & - & 13.89 & .8521 & .1467 & 16.48 & .7860 & .2726 & 12.92 & .8112 & .1398 & 13.03 & .7470 & .1647 \\
        
        \midrule
        \multirow{3}{*}{\makecell[c]{ERGB-Based \\ Deblurring \\ HDR NVS}}
        & HDR-NeRF+       & Blurry ($\Delta t_2$) & \Checkmark & 17.93 & .5986 & .3015 & 17.90 & .8002 & .2967 & 17.77 & .7607 & .2310 & 15.45 & .7002 & .3215 \\
        & EvHDR-NeRF~\cite{Evhdr-NeRF}      & Blurry ($\Delta t_2$) & \Checkmark & 20.63 & .8312 & .2983 & 20.91 & .7667 & .4215 & 21.24 & .8450 & .3948 & 23.93 & .8315 & .3056 \\
        
        & See-NeRF        & Blurry ($\Delta t_2$) & \Checkmark & \textbf{23.55} & \textbf{.9117} & \textbf{.1447} & \textbf{26.74} & \textbf{.9102} & \textbf{.2127} & \textbf{26.01} & \textbf{.9357} & \textbf{.1604} & \textbf{24.13} & \textbf{.8746} & \textbf{.1850} \\

        \bottomrule
    \end{tabular}
    \vspace{-1em}
\end{table*}

\begin{table*}[t]
    \centering
    \caption{Detailed quantitative novel exposure NVS results on each scene of our real dataset.}
    \label{tab:sup8}
    \scriptsize
    \setlength{\tabcolsep}{0.35mm}
        \begin{tabular}{l|cc|ccc|ccc|ccc|ccc|ccc}
        
        \toprule
        & \multicolumn{2}{c|}{Input} 
        & \multicolumn{3}{c}{Bear} 
        & \multicolumn{3}{|c}{Computer} 
        & \multicolumn{3}{|c}{Computerhard} 
        & \multicolumn{3}{|c}{Desktop}
        & \multicolumn{3}{|c}{Table}\\
        
        \midrule
        Methods & Image & Event & PSNR\textuparrow & SSIM\textuparrow & LPIPS\textdownarrow & PSNR\textuparrow & SSIM\textuparrow & LPIPS\textdownarrow & PSNR\textuparrow & SSIM\textuparrow & LPIPS\textdownarrow & PSNR\textuparrow & SSIM\textuparrow & LPIPS\textdownarrow & PSNR\textuparrow & SSIM\textuparrow & LPIPS\textdownarrow \\
        
        \midrule
        HDR-NeRF\textsuperscript{ref} & Sharp ($\Delta t_0$, $\Delta t_2$, $\Delta t_4$) & - & 32.89 & .9611 & .1208 & 28.95 & .9451 & .0675 & 32.69 & .9685 & .0440 & 35.71 & .9612 & .1353 & 36.97 & .9732 & .1072  \\

        \midrule
        HDR-NeRF        & Sharp ($\Delta t_2$) & - & 20.05 & .9138 & .1498 & 20.95 & .8721 & .1098 & 20.44 & .8787 & .0977 & 24.65 & .9102 & .1381 & 24.82 & .9315 & .1196 \\
        HDR-GS          & Sharp ($\Delta t_2$) & - & 18.35 & .8736 & .1901 & 18.76 & .7913 & .1863 & 19.19 & .8257 & .1537 & 21.03 & .8965 & .2218 & 18.83 & .8163 & .2382 \\
        Gaussian-DK & Sharp ($\Delta t_2$) & - & 25.32 & .9541 & .1201 & 23.50 & .8821 & .1085 & 21.83 & .7411 & .1257 & 30.79 & .9440 & .1348 & 30.61 & .9619 & .1030 \\
        GaussHDR          & Sharp ($\Delta t_2$) & - & 25.07 & .9454 & .1862 & 24.63 & .9236 & .1777 & 23.28 & .9101 & .1763 & 30.13 & .9454 & .1829 & 30.74 & .9561 & .1626 \\
        
        \midrule
        HDR-NeRF+       & Blurry ($\Delta t_2$) & \Checkmark & 17.13 & .7716 & .3123 & 13.30 & .5075 & .2917 & 12.60 & .3793 & .4650 & 16.93 & .6431 & .3710 & 16.55 & .6665 & .2819 \\
        EvHDR-NeRF      & Blurry ($\Delta t_2$) & \Checkmark & 22.14 & .9059 & .1937 & 23.91 & .8760 & .1228 & 21.12 & .8187 & .1598 & 25.94 & .8984 & .1777 & 25.68 & .9160 & .1709 \\
        
        See-NeRF        & Blurry ($\Delta t_2$) & \Checkmark & \textbf{30.24} & \textbf{.9591} & \textbf{.1202} & \textbf{27.04} & \textbf{.9481} & \textbf{.0713} & \textbf{24.15} & \textbf{.8576} & \textbf{.1154} & \textbf{31.84} & \textbf{.9536} & \textbf{.1155} & \textbf{32.27} & \textbf{.9621} & \textbf{.1076}  \\
        \bottomrule
    \end{tabular}
    \vspace{-1em}
\end{table*}

\begin{table*}[t]
    \centering
    \caption{Detailed quantitative HDR NVS results on each scene of our real dataset.}
    \label{tab:sup9}
    \scriptsize
    \setlength{\tabcolsep}{0.35mm}
        \begin{tabular}{l|cc|ccc|ccc|ccc|ccc|ccc}
        
        \toprule
        & \multicolumn{2}{c|}{Input} 
        & \multicolumn{3}{c}{Bear} 
        & \multicolumn{3}{|c}{Computer} 
        & \multicolumn{3}{|c}{Computerhard} 
        & \multicolumn{3}{|c}{Desktop}
        & \multicolumn{3}{|c}{Table}\\
        
        \midrule
        Methods & Image & Event & PSNR\textuparrow & SSIM\textuparrow & LPIPS\textdownarrow & PSNR\textuparrow & SSIM\textuparrow & LPIPS\textdownarrow & PSNR\textuparrow & SSIM\textuparrow & LPIPS\textdownarrow & PSNR\textuparrow & SSIM\textuparrow & LPIPS\textdownarrow & PSNR\textuparrow & SSIM\textuparrow & LPIPS\textdownarrow \\
        
        \midrule
        HDR-NeRF\textsuperscript{ref} & Sharp ($\Delta t_0$, $\Delta t_2$, $\Delta t_4$) & - & 30.79 & .9479 & .1045 & 28.08 & .8976 & .1226 & 17.03 & .6914 & .2024 & 27.15 & .9006 & .2036 & 28.23 & .9207 & .1550 \\

        \midrule
        HDR-NeRF        & Sharp ($\Delta t_2$) & - & 10.15 & .4789 & .4333 & 10.28 & .4684 & .4552 & 9.76 & .4989 & .4850 & 11.71 & .6101 & .3860 & 12.47 & .6804 & .3551 \\
        HDR-GS          & Sharp ($\Delta t_2$) & - & 17.11 & .8773 & .3282 & 14.29 & .5777 & .5406 & 9.25 & .4036 & .6376 & 15.00 & .7993 & .4103 & 15.24 & .8140 & .4334 \\
        Gaussian-DK & Sharp ($\Delta t_2$) & - & 14.27 & .8181 & .3253 & 12.75 & .6314 & .4308 & 11.09 & .4588 & .5935 & 16.47 & .8113 & .3343 & 17.08 & .8367 & .3436 \\
        GaussHDR          & Sharp ($\Delta t_2$) & - & 19.55 & .9288 & .1939 & 16.35 & .8305 & .2451 & 17.40 & .7771 & .1689 & 24.21 & .8934 & .1432 & 24.43 & .9047 & .1102 \\
        
        \midrule
        HDR-NeRF+       & Blurry ($\Delta t_2$) & \Checkmark & 21.10 & .8996 & .2196 & 18.09 & .7855 & .2772 & 12.18 & .6128 & .4355 & 18.68 & .8443 & .3110 & 23.01 & .9039 & .1925 \\
        EvHDR-NeRF      & Blurry ($\Delta t_2$) & \Checkmark & 18.17 & .7972 & .2138 & 23.18 & .8251 & .2216 & 10.39 & .1939 & .4011 & 21.85 & .8000 & .2465 & 21.40 & .8649 & .2229 \\
        
        See-NeRF        & Blurry ($\Delta t_2$) & \Checkmark & \textbf{31.91} & \textbf{.9558} & \textbf{.0999} & \textbf{25.85} & \textbf{.8982} & \textbf{.1379} & \textbf{19.43} & \textbf{.7979} & \textbf{.2469} & \textbf{28.13} & \textbf{.9152} & \textbf{.1796} & \textbf{27.13} & \textbf{.9093} & \textbf{.1548} \\
        \bottomrule
    \end{tabular}
    \vspace{-1em}
\end{table*}

\begin{table*}
    \centering
    \caption{Detailed quantitative deblurring NVS results on each scene of the Real-World-Challenge dataset.}
    \label{tab:sup10}
    \scriptsize
    \setlength{\tabcolsep}{0.6mm}
    
    \begin{tabular}{c|c|ccc|ccc|ccc|ccc|ccc}
    \toprule
     & & \multicolumn{3}{c}{Corridor} & \multicolumn{3}{c}{Lab}  & \multicolumn{3}{c}{Lobby} & \multicolumn{3}{c}{Shelf} & \multicolumn{3}{c}{Table}  \\
     & Methods & PSNR\textuparrow  & SSIM\textuparrow  & LPIPS\textdownarrow  & PSNR\textuparrow & SSIM\textuparrow & LPIPS\textdownarrow    & PSNR\textuparrow & SSIM\textuparrow & LPIPS\textdownarrow    & PSNR\textuparrow & SSIM\textuparrow     & LPIPS\textdownarrow  & PSNR\textuparrow & SSIM\textuparrow & LPIPS\textdownarrow   \\

    \midrule
    \multirow{2}{*}{\makecell[c]{RGB-Based \\ Deblurring NVS}}
    &BAD-NeRF & 26.24 & .8950 & .4761 & 28.86 & .9125 & .2386 & 23.06 & .8175 & .6265 & 29.61 & .8820 & .2487 & 20.39 & .7804 & .6102 \\
    &DP-NeRF & 29.64 & .9508 & .3361 & 31.40 & .9452 & .2350 & 27.52 & .9168 & .3093 & 29.92 & .9036 & .2937 & 25.79 & .8964 & .3336 \\
    
    \midrule
    \multirow{5}{*}{\makecell[c]{ERGB-Based \\ Deblurring NVS}}
    & EBAD-NeRF  & 28.74 & .9288 & .4018 & 30.10 & .9354 & .2052 & 26.66 & .8867 & .4048 & 29.88 & .8965 & .2646 & 23.34 & .8574 & .3981 \\

    & Ev-DeblurNeRF & 31.27 & .9546 & .2960 & 31.64 & .9570 & .1790 & 22.48 & .8611 & .4917 & 31.53 & .9285 & .2141 & 22.23 & .8647 & .3975 \\
    & E2GS & 23.83 & .8699 & .5516 & 32.84 & .9571 & .2131 & 27.92 & .9132 & .4155 & 31.81 & .9205 & .2348 & 26.91 & .9218 & .2807 \\
    & E\textsuperscript{2}NeRF & 30.13 & .9584 & .2228 & 32.16 & .9581 & .1717 & 28.94 & .9258 & .2635 & 31.23 & .9109 & .2578 & 27.15 & .9196 & .2622  \\
    & E\textsuperscript{3}NeRF & 31.50 & .9636 & .2213 & 34.02 & .9641 & .1505 & 30.25 & .9339 & .2479 & 32.48 & .9342 & .1736 & \textbf{28.75} & .9362 & .2067  \\

    \midrule
    \multirow{2}{*}{\makecell[c]{ERGB-Based \\ Deblurring HDR NVS}}
    & EvHDR-NeRF  & 29.98 & .9444 & .3701 & 27.34 & .9078 & .3400 & 27.24 & .9064 & .3544 & 28.08 & .8639 & .3345 & 23.32 & .8575 & .4665 \\
        
    & See-NeRF                & \textbf{34.12} & \textbf{.9749} & \textbf{.1313} & \textbf{35.21} & \textbf{.9712} & \textbf{.1277} & \textbf{31.12} & \textbf{.9449} & \textbf{.2248} & \textbf{34.40} & \textbf{.9545} & \textbf{.1065} & 28.64 & \textbf{.9367} & \textbf{.1965}  \\
    \bottomrule
    \end{tabular}
    \vspace{-1em}
\end{table*}

\begin{figure*}[t]
  \centering
  \includegraphics[width=1\linewidth]{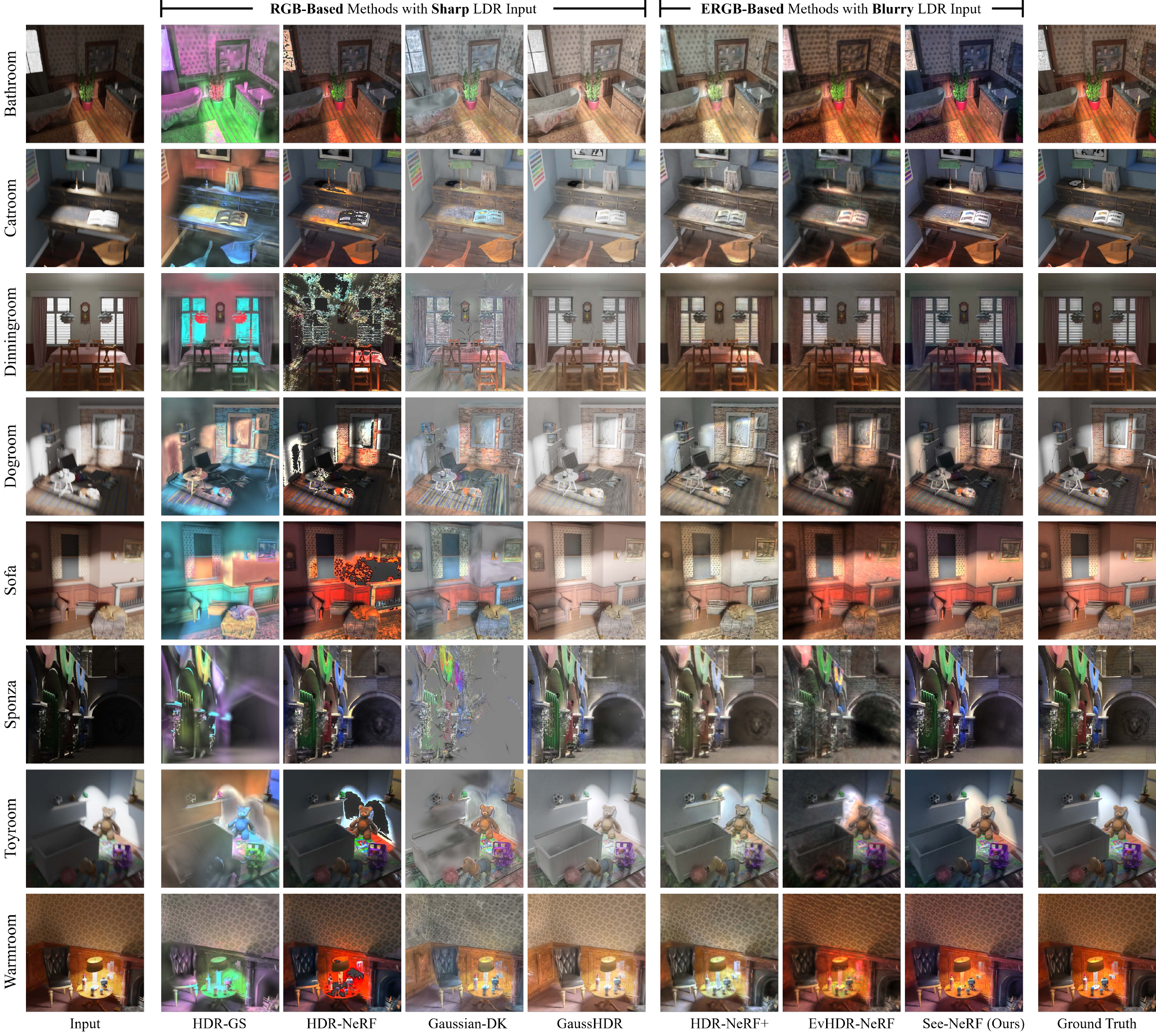}
   \caption{Detailed qualitative HDR NVS results on each scene of our synthetic dataset.}
   \label{fig:sup4}
   \vspace{-1em}
\end{figure*}

\begin{figure*}[t]
  \centering
  \includegraphics[width=1\linewidth]{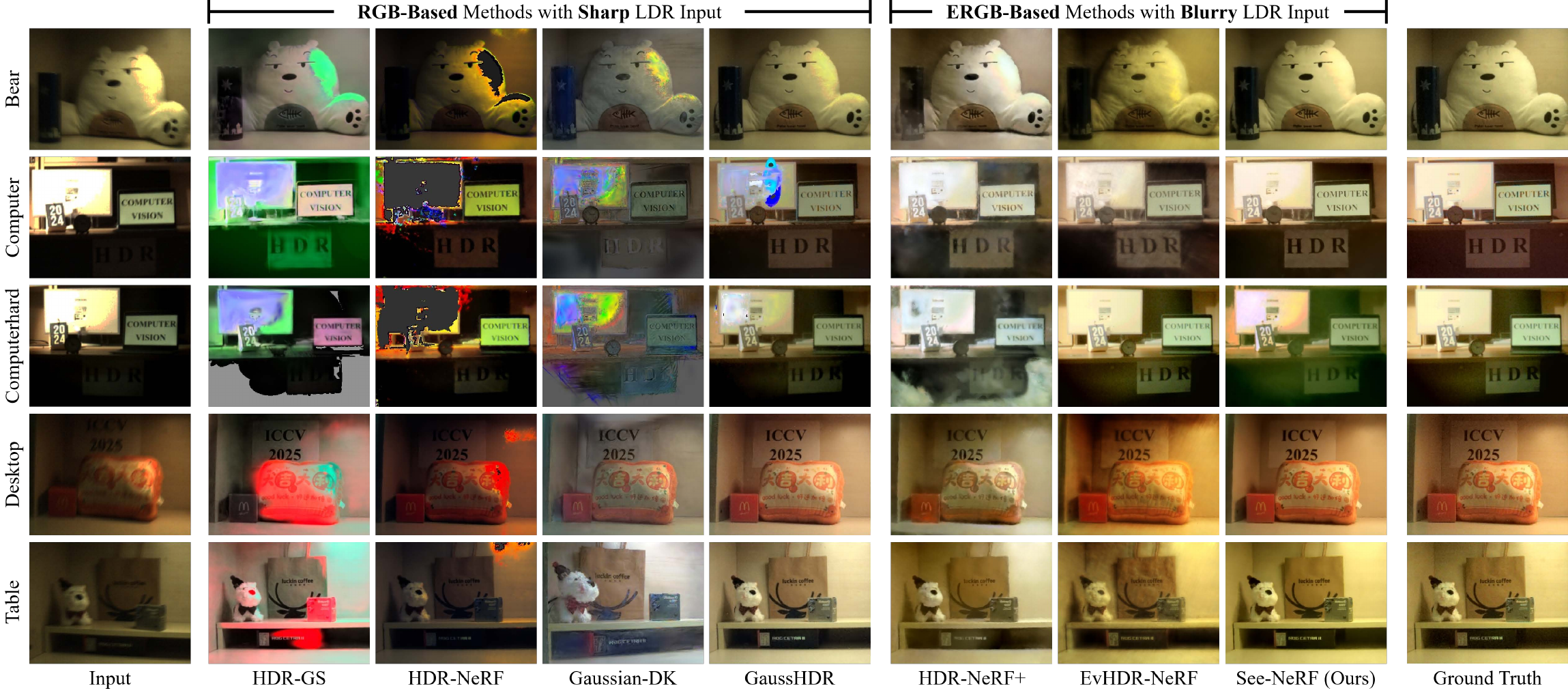}
   \caption{Detailed qualitative HDR NVS results on each scene of our real dataset.}
   \label{fig:sup5}
\end{figure*}

\begin{figure*}[t]
  \centering
  \includegraphics[width=1\linewidth]{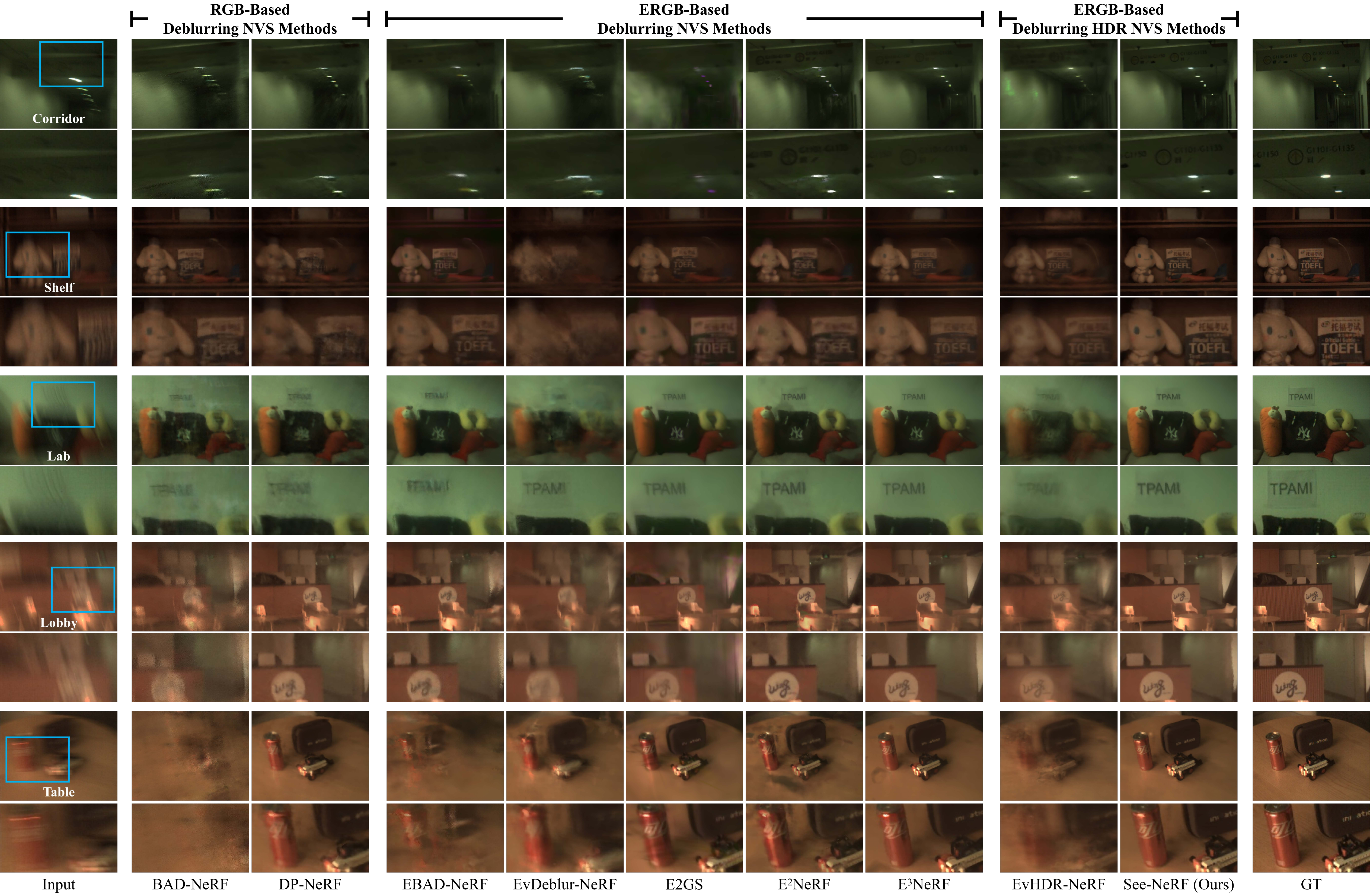}
   \caption{Detailed qualitative deblurring NVS results on each scene of the Real-World-Challenge dataset.}
   \label{fig:sup6}
\end{figure*}

\section{Supplementary detailed results}
\label{sec:E}

\subsection{Detailed quantitative results}
\label{sec:E.1}

\myPara{Our synthetic dataset:}
\cref{tab:sup4} and \cref{tab:sup5} illustrate the detailed quantitative novel exposure NVS results on each scene of our synthetic dataset.
See-NeRF achieves the best overall performance.
In certain scenes, the novel exposure results of the compared RGB-based HDR NVS methods are better than those of See-NeRF because it uses sharp images as input.
Besides, the detailed quantitative results in \cref{tab:sup6} and \cref{tab:sup7} illustrate that their performance drops significantly on the HDR NVS task, indicating that they are overfitted to the LDR image supervision, thus learning a wrong scene HDR representation.
In contrast, See-NeRF achieves the best HDR NVS results for all three metrics.

\myPara{Our real dataset:}
\cref{tab:sup8} and \cref{tab:sup9} illustrate the detailed quantitative novel exposure and HDR NVS results on each scene of our real dataset.
Some results of the compared RGB-based HDR NVS methods are better than those of See-NeRF on LPIPS due to the sharp images input, while our See-NeRF still achieves the best overall performance.
For the HDR experiment, See-NeRF is even better than HDR-NeRF\textsuperscript{ref} in ``Bear'', ``Computerhard'', and ``Desktop'' scenes, resulting in its average results in Tab. 1 of the main paper outperforming HDR-NeRF\textsuperscript{ref}.
This is likely attributable to the low imaging quality of the DAVIS 346 RGB sensor, which is not suitable for RGB-based methods.

\myPara{Real-World-Challenge dataset:}
\cref{tab:sup10} illustrates the detailed quantitative deblurring NVS results on each scene of the Real-World-Challenge dataset~\cite{e3nerf}.
Our See-NeRF realizes the SOTA performance compared to the previous RGB-based and ERGB-based deblurring NVS methods.

\subsection{Detailed qualitative results}
\label{sec:E.2}

\myPara{Our synthetic dataset:}
\cref{fig:sup4} shows the detailed qualitative HDR NVS results on our synthetic dataset.
The results of See-NeRF are closest to GT across all scenes.
HDR-GS suffers from severe color shifts.
HDR-NeRF fails to reconstruct the overexposed and underexposed areas.
HDR-NeRF+ is limited by the HDREv-Net~\cite{evhdr-video2}, which only achieves a brightening effect relative to the input image without fully restoring the scene color details.
EvHDR-NeRF still suffers from slight color casts and motion blur.

\myPara{Our real dataset}
\cref{fig:sup5} shows the detailed qualitative HDR NVS results on our real dataset, and the results of See-NeRF are closest to GT across all scenes.
The compared methods show the same phenomenon as in the synthetic data.

\myPara{Real-World-Challenge dataset}
\cref{fig:sup6} shows the detailed qualitative deblurring NVS results on the Real-World-Challenge dataset.
See-NeRF shows impressive deblurring performance compared to other methods.
Note that EvHDR-NeRF is significantly affected by the motion blur.

\subsection{Supplementary video results}
\label{sec:E.3}
We also provide supplementary video rendering results to show the impressive HDR and deblurring novel view synthesis results of See-NeRF and the robustness to noise events of our framework.
We highly recommend that the reviewers refer to the attached video for details on our project web page: \url{https://icvteam.github.io/See-NeRF.html}.
 
\end{document}